\documentclass{article} % For LaTeX2e
\usepackage{iclr2027_conference,times}

\usepackage{amsmath,amsfonts,bm}

\def\eqref#1{equation~\ref{#1}}
\def\1{\bm{1}}

\def\va{{\bm{a}}}

\def\vu{{\bm{u}}}

\def\vx{{\bm{x}}}
\def\vy{{\bm{y}}}
\def\vz{{\bm{z}}}

\DeclareMathAlphabet{\mathsfit}{\encodingdefault}{\sfdefault}{m}{sl}
\SetMathAlphabet{\mathsfit}{bold}{\encodingdefault}{\sfdefault}{bx}{n}

\usepackage{hyperref}
\usepackage{url}

\usepackage{booktabs}
\usepackage{multirow}
\usepackage{makecell}
\usepackage{graphicx}
\usepackage{lipsum}
\usepackage{graphicx}
\usepackage{wrapfig}
\usepackage{algorithm}
\usepackage{algpseudocode}
\usepackage{amsthm, amssymb}
\usepackage{thmtools,thm-restate}
\usepackage{colortbl}

\usepackage{caption}
\usepackage{amssymb}
\usepackage{pifont}

\usepackage{tabularx}
\usepackage{array}

\def\eqref#1{Eq.~(\ref{#1})}

\newcommand{\cmark}{\checkmark}
\newcommand{\xmark}{\ding{55}}

\newtheorem{lemma}{Lemma}

\title{Learning Noise Optimization for Reward Alignment in Few-Step Generative Models}

\author{%
  Hongeun Kim, Jeongsol Kim, Jong Chul Ye\\
  KAIST AI\\
  \texttt{\{hongeun, jeongsol, jong.ye\}@kaist.ac.kr}\\
  \text{* Equal contribution}\\
}

\iclrfinalcopy % Uncomment for camera-ready version, but NOT for submission.

\fancypagestyle{plain}{%
  \fancyhf{}
  \fancyhead[L]{\small{Preprint.}}
  \renewcommand{\headrulewidth}{0.4pt}
  \fancyfoot[c]{\rm\thepage} % page number
}

\begin{document}

\maketitle

\begin{figure}[!h]
    \centering
    \includegraphics[width=1.0\linewidth,height=8.0cm]{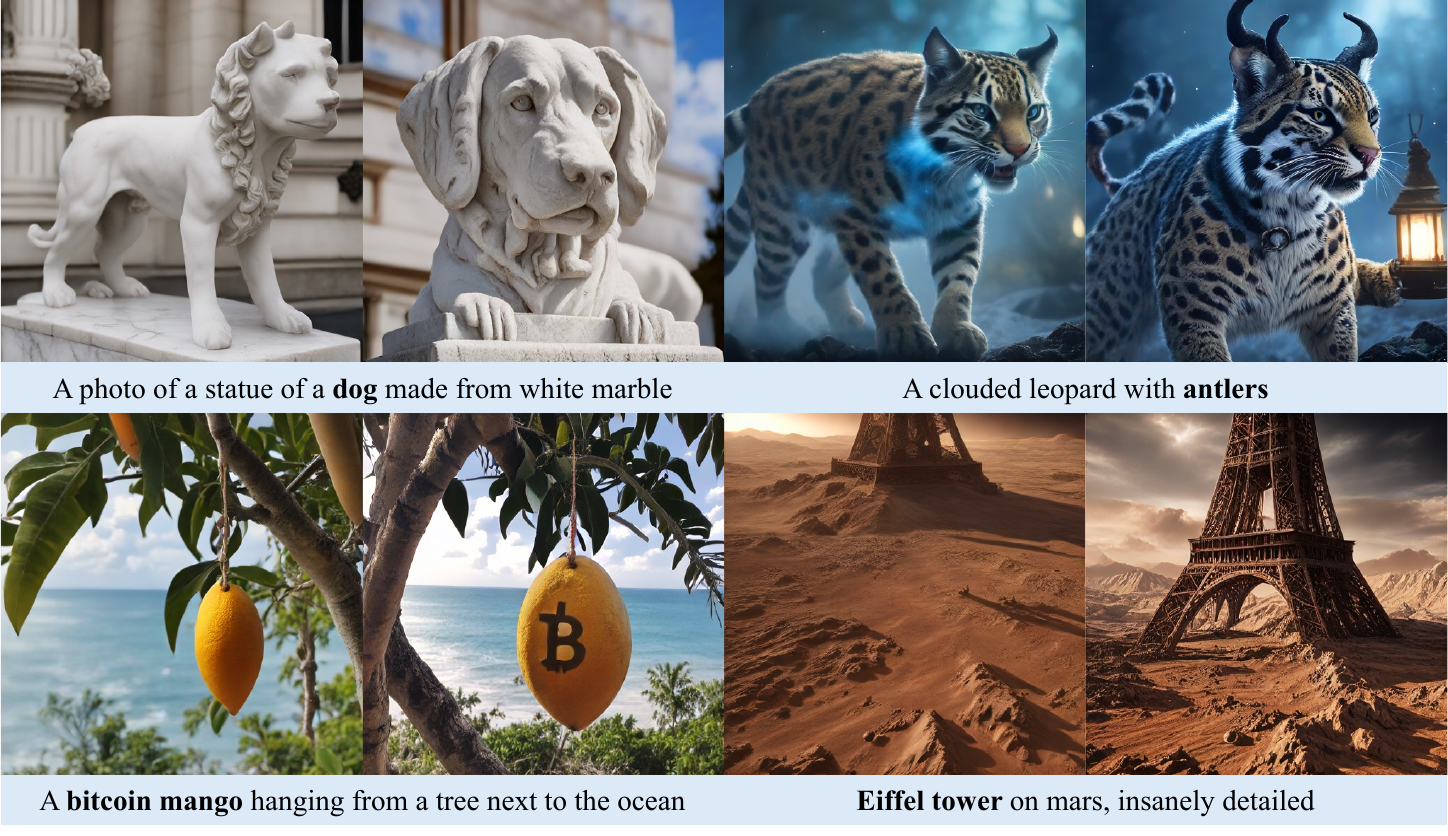}
    \vspace{-4mm}
    \caption{LeNO learns reward-improving corrections directly in initial noise space.
    Each pair shows generations before (left) and after (right) applying the
    learned noise correction, with the pretrained generator kept frozen. The corrected noises produce images that better capture the desired visual attributes across diverse prompts.}
    \label{fig:main}
\end{figure}

\begin{abstract}
Reward alignment for diffusion- and flow-based generative models is often formulated over stochastic trajectories, making it less suitable for few-step deterministic generators.
Initial noise optimization provides a trajectory-free alternative, but existing gradient-free methods require costly per-sample search at inference.
We introduce \textbf{LeNO} (\textbf{Le}arning \textbf{N}oise \textbf{O}ptimization), which amortizes reward-guided initial noise optimization into a learned residual transport while keeping the pretrained generator frozen.
LeNO learns an anchor-conditioned distribution over noise corrections using Local GRPO and scalar rewards, eliminating iterative test-time optimization.
For Gaussian initial noise with norm-preserving projection, we derive a tractable upper bound on the KL divergence between the corrected noise distribution and the original prior, providing a principled regularization objective.
We evaluate LeNO on few-step text-to-image and protein structure generation, including expensive black-box designability rewards.
Across both domains, LeNO improves target rewards with a single noise correction at inference while largely preserving generation quality and diversity.
\end{abstract}

\section{Introduction}

Reward-based post-training has emerged as a powerful paradigm for adapting pretrained generative models toward objectives such as human preference, visual quality, and task-specific utility. In diffusion and flow models, many recent approaches formulate generation as a sequential decision process and optimize the generative trajectory using reinforcement learning or stochastic control~\citep{black2024training,liu2026flowgrpo,lee2025pcpo}.
Recent generative models, however, increasingly rely on few-step or even one-step deterministic sampling~\citep{song2023consistency, yin2024one, sauer2024adversarial, geng2025mean}. While highly efficient, these models provide only few step deterministic generation trajectory, making conventional trajectory-level policy optimization less natural.

A promising alternative is to move reward optimization from the generative trajectory to the \emph{initial noise space}, as the initial noise strongly influences the generated output. This has motivated recent inference time methods that search for reward-favorable noises while keeping the pretrained generator frozen~\citep{eyring2024reno,yoon2026psisampler,kim2026gradient}.
While these methods enable deterministic generators to be aligned with a given reward, including non-differentiable functions, they require an iterative noise-update process involving repeated generator executions and reward evaluations for each test sample.
This becomes particularly costly when either generation or reward evaluation requires an expensive external pipeline.

This motivates \emph{amortizing} reward-guided noise optimization into a learned mapping.
Existing approaches, however, typically require either pre-collected offline data or differentiable reward functions. For example, GoldenNoise~\citep{zhou2025golden} learns from a pre-constructed dataset of favorable noise--sample pairs, while  HyperNoise~\citep{eyring2026noise} learns to modify the initial noise through differentiable reward feedback. In contrast, directly learning sample-dependent noise corrections from online scalar rewards, without offline data or reward gradients, remains relatively underexplored.

To address these limitations, we introduce \textbf{LeNO}
(\textbf{Le}arning \textbf{N}oise \textbf{O}ptimization), a framework that
amortizes per-sample initial noise optimization for deterministic
generators under differentiable or non-differentiable rewards.
Rather than predicting an optimized noise sample from scratch, LeNO
parameterizes the mapping in residual form and learns an
anchor-dependent correction to the initial noise.
To optimize this residual mapping using only scalar rewards, we formulate
it as a stochastic transport that samples Gaussian residual actions around
the predicted correction mean.
These actions are trained with Local GRPO, where group-relative reward
comparisons among corrections sampled from the same anchor provide the
learning signal.
We further provide a distributional interpretation of this residual
parameterization.
For Gaussian initial noise with anchor-norm-preserving projection, we
show that a scaled squared norm of the predicted residual mean
upper-bounds the KL divergence between the corrected noise distribution
and the prior.
This provides a tractable lower bound on the KL-regularized reward
objective and directly motivates the residual regularization in Local GRPO.
At inference, the learned transport generates a corrected initial noise
with a single residual sample, eliminating per-sample optimization and
additional reward evaluations.

We evaluate LeNO on few-step text-to-image generation and protein structure generation. 
In text-to-image generation, we consider direct optimization of multiple preference rewards. 
For proteins, we optimize a black-box designability objective whose
evaluation requires sequence design and structure prediction, a setting where
gradient-based amortization is inapplicable and per-sample search is costly.
Across these settings, LeNO improves target rewards with only a lightweight residual correction at inference time, while retaining favorable held-out compositional performance and structural diversity. 
These results demonstrate that learned initial noise correction provides an efficient and broadly applicable alternative to repeated per-sample reward optimization.

\section{Background}

\textbf{Reward-Tilted Noise Distribution.}
For a deterministic generator $\vx=G_\phi(\vz,c)$ with initial noise
$\vz\sim p_0(\vz)$, the generated sample is fully determined by $z$ once the
condition $c$ is fixed. Let $r(\cdot)$ denote a scalar, potentially black-box reward function defined on the generated sample. Reward-based alignment can therefore be formulated
directly in the initial noise space by tilting the noise prior toward high-reward regions:
\vspace{-1mm}
\begin{equation}
p^\star(\vz\mid c)
\propto
p_0(\vz)
\exp\left(
\frac{r(G_\phi(\vz,c))}{\lambda}
\right),
\end{equation}
where $\lambda>0$ controls the trade-off between reward maximization and
preservation of the original noise prior. This formulation provides a
direct basis for initial noise optimization, where the pretrained generator
is kept fixed and reward alignment is performed by searching for
higher-reward noise samples.

\textbf{Zeroth-Order Initial Noise Optimization.}
Reward optimization can be performed directly in the initial noise space
without differentiating through either the generator or the reward function.
Let
\begin{equation}
\mathcal{R}(\vz;c)
:=
r\!\left(G_\phi(\vz,c)\right),
\qquad
\mathcal{R}_\sigma(\vz;c)
:=
\mathbb{E}_{\epsilon\sim\mathcal{N}(0,I)}
\left[
\mathcal{R}(\vz+\sigma\epsilon;c)
\right].
\end{equation}
The gradient $g(\vz,c)$ of the Gaussian-smoothed reward can be estimated
from centered reward-weighted perturbations as
\vspace{-2mm}
\begin{equation}
\label{eq:zo_estimator}
g(\vz,c)
:=
\nabla_{\vz}\mathcal{R}_{\sigma}(\vz;c)
\approx
\frac{1}{(K-1)\sigma}
\sum_{i=1}^{K}
\left(
R_i-\bar R_{\vz}
\right)\epsilon_i ,
\end{equation}
where
$R_i=\mathcal{R}(\vz+\sigma\epsilon_i;c)$ and
$\bar R_{\vz}=\frac{1}{K}\sum_{i=1}^{K}R_i$.
This form underlies Natural Evolution Strategy style~\citep{JMLR:v15:wierstra14a} gradient estimation and related gradient-free initial noise optimization methods, including approaches that derive reward-weighted perturbation updates from stochastic control
perspectives~\citep{kim2026gradient}. Such methods avoid reward
backpropagation, but require estimating a new local direction for each noise
state through repeated generator and reward evaluations. LeNO instead seeks to amortize these state-dependent noise corrections into a learned residual transport, avoiding repeated optimization at inference time.

\textbf{Policy Optimization for Deterministic Generative Models.}
Recent approaches have explored how to enable policy optimization for
deterministic generative models by recovering a suitable source of
stochasticity. Flow-GRPO~\citep{liu2026flowgrpo}, for example, converts the
deterministic probability-flow ODE into an equivalent stochastic differential
equation, thereby introducing stochastic transitions on which policy-gradient
optimization can be performed. An alternative perspective preserves the
deterministic sampling dynamics and instead constructs stochasticity through
local trajectory neighborhoods.

Neighbor GRPO~\citep{he2026neighbor} follows this local-neighborhood
perspective. Given a shared reference noise $\epsilon^\star$, it samples
neighboring initial conditions
\begin{equation}
\epsilon^{(i)}
=
\sqrt{1-\sigma^2}\epsilon^\star
+
\sigma\delta^{(i)},
\qquad
\delta^{(i)}\sim\mathcal{N}(0,I),
\end{equation}
and propagates them through the same deterministic ODE to construct a local
trajectory neighborhood. Despite the similar shared-anchor perturbation geometry, Neighbor GRPO optimizes the generative model through a contrastive, distance-based surrogate policy defined over neighboring ODE trajectories. In contrast, LeNO keeps the
generator frozen and learns a separate stochastic residual transport in the
initial noise space, using local perturbations to obtain reward-based
training signals.
The two approaches are therefore complementary, as LeNO leaves
the pretrained model unchanged and applies to any generator that
exposes an initial noise input.

\section{Methods}

\subsection{Noise-Space Residual Transport}
\label{subsec:noise_space_transport}

We formulate amortized initial noise optimization as a stochastic
residual transport for a frozen pretrained generator $G_\phi$.
Given an initial noise $\vz\sim p_0$ and condition $c$, we learn
an anchor-dependent residual mean $\mu_\theta(\vz,c)$ to improve
the generated output using scalar reward feedback.
The transport samples a residual action and applies it to the anchor:
\begin{align}
\va
&=
\eta\mu_\theta(\vz,c)+\sigma\epsilon,
\qquad \epsilon\sim\mathcal{N}(0,I),
\label{eq:exploration_distribution}
\\
\vz'
&=
F_\theta(\vz,\epsilon,c)
:=
\mathcal{T}(\vz,\va),
\label{eq:residual_map}
\end{align}
where $\eta>0$ controls the correction scale and $\sigma>0$
is a fixed sampling scale. For Gaussian initial noise, we use the norm-preserving transition
\begin{equation}
\mathcal{T}(\vz,\va)
=
\operatorname{NormProj}(\vz+\va;\|\vz\|_2),
\qquad
\operatorname{NormProj}(\vx;r)
:=
r\,\frac{\vx}{\|\vx\|_2}.
\label{eq:norm_projection}
\end{equation}
This updates the direction of the initial noise while preserving
its norm. The stochastic transport induces the corrected noise distribution
\begin{equation}
\vz' = F_\theta(\vz,\epsilon,c)
\sim q_\theta(\cdot\mid c),
\label{eq:pushforward_distribution}
\end{equation}
where $\vz\sim p_0$ and $\epsilon\sim\mathcal{N}(0,I)$
are independent.

At inference, we sample one corrected noise and generate
$G_\phi(\vz',c)$.
During training, we draw multiple actions per anchor and compare
their rewards to learn the residual mean.
The same conditional Gaussian structure provides both a
tractable distributional KL bound and an action likelihood
for reward-based learning.
We next derive the bound, followed by the Local GRPO objective.

\subsection{Distributional Target and KL Regularization}
\label{subsec:distributional_kl}

We aim to align the corrected noise distribution
$q_\theta(\cdot\mid c)$ with the reward-tilted target
\begin{equation}
p^\star(\vz\mid c)
=
\frac{
p_0(\vz)\exp(\mathcal{R}(\vz;c)/\lambda)
}{
Z(c)
},
\qquad \lambda>0,
\label{eq:reward_tilted_pushforward}
\end{equation}
where $p_0=\mathcal{N}(0,I)$, $\mathcal{R}(\vz;c)=r(G_\phi(\vz,c))$ and
$Z(c):=\mathbb{E}_{\vz\sim p_0}
[\exp(\mathcal{R}(\vz;c)/\lambda)]<\infty$.
This target favors high-reward noises while retaining
the prior as a reference.

Whenever the relevant expectations and divergences are finite,
\begin{equation}
\begin{aligned}
D_{\mathrm{KL}}\!\left(
q_\theta(\cdot\mid c)\Vert p^\star(\cdot\mid c)
\right)
&=
D_{\mathrm{KL}}\!\left(
q_\theta(\cdot\mid c)\Vert p_0
\right)
\\
&\quad-
\frac{1}{\lambda}
\mathbb{E}_{\vz\sim q_\theta(\cdot\mid c)}
[\mathcal{R}(\vz;c)]
+\log Z(c).
\end{aligned}
\label{eq:reward_tilted_objective}
\end{equation}
Since $Z(c)$ is independent of $\theta$, minimizing this
KL divergence is equivalent to maximizing
\begin{equation}
\mathcal{L}(q_\theta;c)
:=
\mathbb{E}_{\vz\sim q_\theta(\cdot\mid c)}
[\mathcal{R}(\vz;c)]
-
\lambda D_{\mathrm{KL}}\!\left(
q_\theta(\cdot\mid c)\Vert p_0
\right).
\label{eq:pushforward_target}
\end{equation}
The reward term can be estimated from generated samples,
but the prior-relative KL remains intractable because
$q_\theta$ is implicitly defined by the transport.
We therefore bound this KL term using the conditional
Gaussian action distributions.

\begin{restatable}[]{prop}{distributionalkl}
\label{prop:distributional_kl}
Let $p_0=\mathcal{N}(0,I)$ and let $q_\theta$ be a corrected noise distribution induced by the
norm-preserving transition $\mathcal{T}$ in \eqref{eq:norm_projection}.
For $\sigma>0$ and
$\mathbb{E}_{\vz\sim p_0}
\|\mu_\theta(\vz,c)\|_2^2<\infty$,
the following upper bound holds:
\begin{equation}
\begin{aligned}
D_{\mathrm{KL}}\!\left(
q_\theta(\cdot\mid c)\Vert p_0
\right)
&\leq
\mathbb{E}_{\vz\sim p_0}
D_{\mathrm{KL}}\!\left(
\mathcal{N}(\eta\mu_\theta(\vz,c),\sigma^2 I)
\,\Vert\,
\mathcal{N}(0,\sigma^2 I)
\right)
\\
&=
\frac{\eta^2}{2\sigma^2}
\mathbb{E}_{\vz\sim p_0}
\|\mu_\theta(\vz,c)\|_2^2.
\end{aligned}
\label{eq:distributional_kl_bound}
\end{equation}
\end{restatable}

The inequality follows from prior preservation under zero-mean residual actions, together with the data-processing inequality applied to the joint distributions of anchors and residual actions. The equality follows from the closed-form KL divergence between Gaussian distributions with identical covariance. A detailed proof is provided in Appendix~\ref{app:distributional_kl}.

Under the conditions of
Proposition~\ref{prop:distributional_kl}, we obtain a
tractable lower bound on the distributional objective:
\vspace{-2mm}
\begin{equation}
\begin{aligned}
\mathcal{L}_{\mathrm{aux}}(\theta;c)
&:=
\mathbb{E}_{\vz'\sim q_\theta(\cdot\mid c)}
[\mathcal{R}(\vz';c)]
-
\frac{\lambda\eta^2}{2\sigma^2}
\mathbb{E}_{\vz\sim p_0}
\|\mu_\theta(\vz,c)\|_2^2
\leq
\mathcal{L}(q_\theta;c).
\end{aligned}
\label{eq:distributional_lower_bound}
\end{equation}
The reward term can be estimated from sampled transport
outputs, while the regularizer depends only on the predicted
residual mean.
Together with \eqref{eq:reward_tilted_objective},
this implies that maximizing $\mathcal{L}_{\mathrm{aux}}$
minimizes an upper bound on the KL divergence from
$q_\theta$ to the reward-tilted target $p^\star$.
Using the same stochastic transport and noise scale during
training and inference ensures that the bound applies to
the corrected noise distribution used for generation.

\subsection{Local GRPO Objective}
\label{subsec:local_grpo}

\noindent\textbf{Training Objective.}
We next optimize the auxiliary reward objective
$\mathcal{L}_{\mathrm{aux}}(\theta;c)$ through a tractable Local GRPO
surrogate.
Rather than maximizing the expected reward directly, Local GRPO uses
group-relative comparisons among samples from the stochastic residual
transport to update the residual mean from scalar rewards, while
preserving the regularization implied by the distributional KL bound.

\begin{wrapfigure}{r}{0.5\textwidth}
\vspace{-4mm}
\begin{minipage}{0.5\textwidth}
\begin{algorithm}[H]
\caption{Training LeNO with Local GRPO}
\label{alg:local_grpo}
\begin{algorithmic}[1]
\Require Frozen generator $G_\phi$, reward $r$, residual transport $F_\theta$,
group size $K$, residual scale $\eta$, exploration scale $\sigma$,
regularization coefficient $\beta$

\For{each training iteration}
    \State $\theta_{\mathrm{old}} \gets \theta$
    \State Sample condition $c$ and anchor $\vz\sim p_0(\vz)$
    \For{$i=1,\ldots,K$}
        \State Sample $\epsilon_i\sim\mathcal{N}(0,I)$
        \State $\va^{(i)}
        \gets\eta\mu_{\theta_{\mathrm{old}}}(\vz,c)+\sigma\epsilon_i$
        \State $\vz^{(i)}\gets\mathcal{T}(\vz,\va^{(i)})$
        \State $R_i\gets r\!\left(G_\phi(\vz^{(i)},c)\right)$
    \EndFor
    \State Compute group-relative advantages $A_i$
    \State Compute exploration likelihood ratios $\rho^{(i)}(\theta)$
    \State Update $\theta$ using \eqref{eq:local_grpo}
\EndFor
\end{algorithmic}
\end{algorithm}
\end{minipage}
\vspace{-5mm}
\end{wrapfigure}

At the start of each training iteration, we snapshot the current
transport parameters as $\theta_{\mathrm{old}}$ and hold them
fixed throughout the subsequent GRPO updates on the sampled
particles. For each anchor $\vz$ and condition $c$, we draw $K$
residual actions from the conditional Gaussian distribution
in \eqref{eq:exploration_distribution}.
For $i=1,\ldots,K,$
\begin{equation}
\va^{(i)}
=
\eta\mu_{\theta_{\mathrm{old}}}(\vz,c)
+
\sigma\epsilon_i,
\label{eq:local_particles}
\end{equation}
where $\epsilon_i\overset{\mathrm{iid}}{\sim}\mathcal{N}(0,I)$.
Each action produces a corrected noise
$\vz^{(i)}=\mathcal{T}(\vz,\va^{(i)})$
with reward $R_i=\mathcal{R}(\vz^{(i)};c)$.
These particles are samples of the same stochastic transport
used at inference, sharing the anchor and predicted residual
mean while differing in their Gaussian perturbations.
We compare their rewards within each group to construct
the group-relative advantages:
\begin{equation}
A_i
=
\frac{R_i-\bar R}{s_R+\varepsilon},
\qquad
\bar R
=
\frac{1}{K}\sum_{j=1}^{K}R_j,
\label{eq:local_advantage}
\end{equation}
where $s_R$ is the within-group reward standard deviation
and $\varepsilon>0$ ensures numerical stability.

Let $\pi_\theta(\va\mid\vz,c)$ denote the Gaussian action
density specified by \eqref{eq:exploration_distribution}.
For the sampled actions, we define the likelihood ratio
\begin{equation}
\rho^{(i)}(\theta)
:=
\frac{
\pi_\theta(\va^{(i)}\mid\vz,c)
}{
\pi_{\theta_{\mathrm{old}}}(\va^{(i)}\mid\vz,c)
}.
\label{eq:policy_ratio}
\end{equation}
This ratio is evaluated on the residual actions before
projection and does not require the density of the corrected
noise marginal $q_\theta$.
Holding the sampled actions and advantages fixed, our penalized
local GRPO maximizes the following:
\begin{equation}
\begin{aligned}
\mathcal{J}(\theta)
=
\mathbb{E}\Bigg[
&\frac{1}{K}\sum_{i=1}^{K}
\min\!\left(
\rho^{(i)}(\theta)A_i,\,
\operatorname{clip}\!\left(
\rho^{(i)}(\theta),
1-\epsilon_{\mathrm{clip}},
1+\epsilon_{\mathrm{clip}}
\right)A_i
\right)
-
\beta\frac{\eta^2}{2\sigma^2}
\|\mu_\theta(\vz,c)\|_2^2
\Bigg],
\end{aligned}
\label{eq:local_grpo}
\end{equation}
where the expectation is over training conditions,
anchors $\vz\sim p_0$, and particles sampled under
$\theta_{\mathrm{old}}$.
Here, $\epsilon_{\mathrm{clip}}>0$ is the PPO clipping
parameter and $\beta>0$ controls regularization strength.
In \eqref{eq:local_grpo},  the first term promotes residual actions with higher
group-relative rewards, while the second regularizes the policy toward
the zero-mean Gaussian reference through the conditional KL divergence,
thereby linking 
 to the distributional KL term in
\eqref{eq:distributional_lower_bound}.
In practice, we combine group-normalized rewards with PPO-style clipping
to obtain a stable Local GRPO surrogate, where $\beta$ controls the
strength of KL regularization relative to the normalized reward scale.

The local sampling procedure also admits a zeroth-order interpretation.
Before norm projection, reward-weighted Gaussian perturbations provide
an estimator of the gradient of the Gaussian-smoothed reward evaluated
at the mean-shifted anchor, closely related to zeroth-order noise
optimization methods such as ZeNO~\citep{kim2026gradient}.
Unlike such methods, which optimize the initial noise independently for
each test-time sample, LeNO uses these local reward signals to learn an
anchor-conditioned transport and thereby amortizes noise optimization
across samples.
See Appendix~\ref{app:zeroth_order_derivation} for more details of the derivation.

\textbf{Why State-Dependent Noise Corrections?}
LeNO conditions the residual mean $\mu_\theta(\vz,c)$ on the anchor $\vz$
itself, rather than solely on the condition $c$.
This design is motivated by the zeroth-order interpretation above:
because local sampling estimates the reward-improvement direction in a
neighborhood of a specific anchor, the resulting direction is generally
anchor-dependent.
In contrast, a correction conditioned only on $c$ cannot adapt to such
local variation and instead captures only an average direction across
anchors.
Appendix~\ref{app:state_dependency_justification} formalizes this
intuition using a simplified first-order surrogate of
$\mathcal{L}_{\mathrm{aux}}$.
Under this surrogate, the optimal anchor-dependent correction follows
the local reward-improvement direction at each anchor, whereas the
optimal anchor-independent correction follows their population average.
The resulting objective gap is proportional to the total variance of
the local directions across anchors and vanishes only when these
directions are identical for all anchors (Lemma~\ref{lemma:state_dependent}).
We empirically confirm the benefit of anchor-dependent conditioning in
Fig.~\ref{fig:state_dependency_ablation}.

\section{Experiments}

\subsection{Toy Experiment}

\begin{figure}
    \centering
    \includegraphics[width=\linewidth]{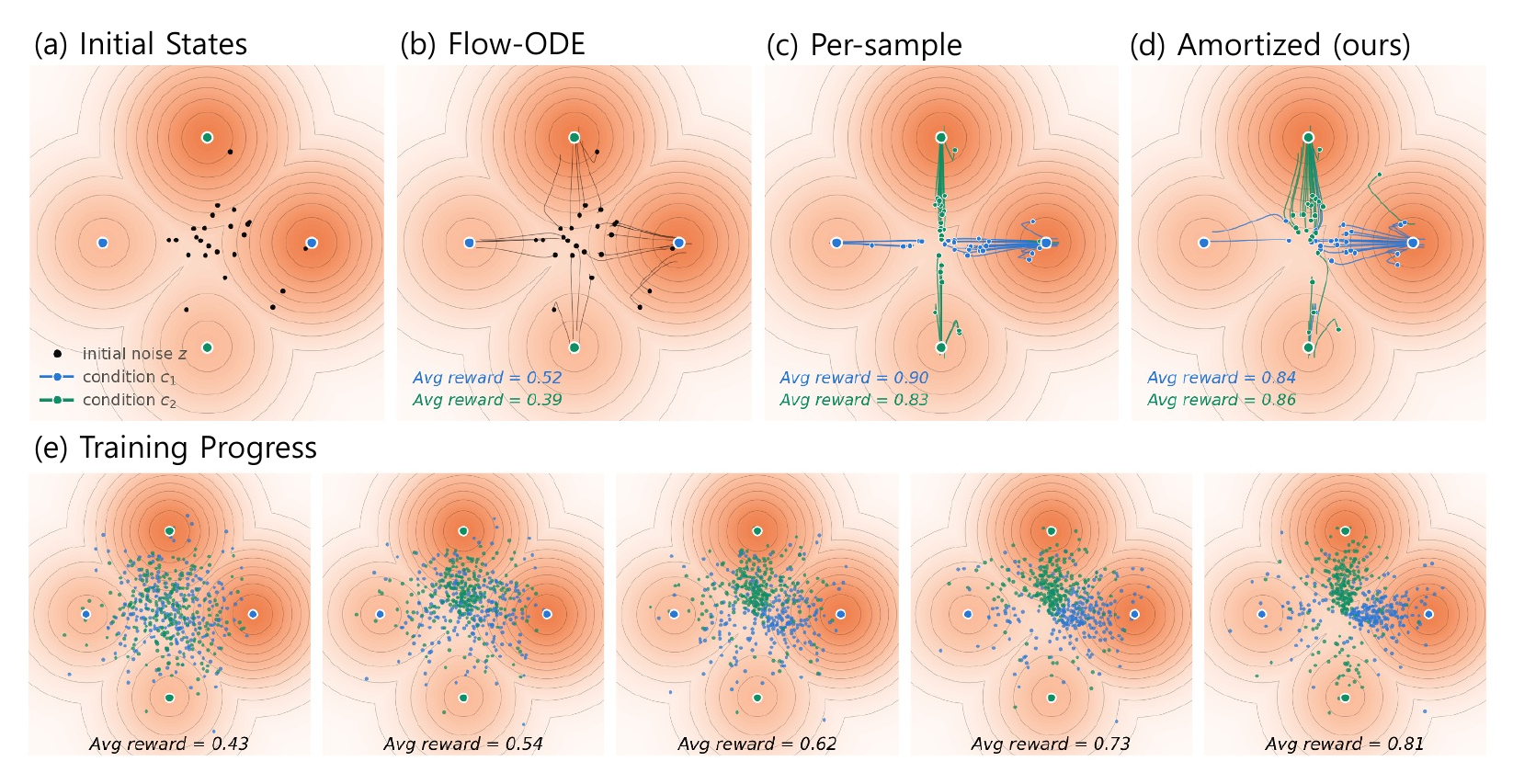}
    \vspace{-4mm}
    \caption{Toy Experiment. 
    (a) Initial noise samples and the reward landscape of a four-component Gaussian mixture.
    (b) Samples obtained by directly passing the initial noise through the frozen probability-flow ODE.
    (c) Per-sample initial noise optimization with 30 step optimization.
    (d) Our amortized noise space correction with the residual transport.
    The learned correction successfully redirects samples toward the condition-specific high-reward modes with substantially lower inference time computation.
    (e) Corrected noise distribution over training.
    }
    \label{fig:toy_figure}
    \vspace{-0.4cm}
\end{figure}

We consider a four-component Gaussian mixture model where two different conditions, $c_1$ and $c_2$, select disjoint pairs of target modes with distinct reward values. The deterministic generator $G_\phi$ is defined as the probability-flow ODE of a linear interpolant between $\mathcal{N}(0,I)$ and the target mixture. Fig.~\ref{fig:toy_figure}(a) illustrates the initial noise samples and the corresponding reward landscape, where darker regions indicate higher rewards. As a baseline, Fig.~\ref{fig:toy_figure}(b) directly maps the initial particles through the frozen flow ODE without any reward-based correction. For per-sample optimization, Fig.~\ref{fig:toy_figure}(c) applies per-sample 30 step optimization, following ZeNO~\citep{kim2026gradient}, with 32 perturbation particles per step, requiring 960 generator evaluations per sample. In contrast, LeNO in Fig.~\ref{fig:toy_figure}(d) uses one-step residual correction with a lightweight 35K-parameter MLP trained for 400 iterations.

Despite replacing iterative per-sample optimization with a learned corrector, our method consistently redirects samples toward the high-reward modes specified by each condition. While per-sample optimization improves rewards for each condition to $0.90$ and $0.83$, our amortized correction attains $0.84$ and $0.86$, showing that condition-dependent reward-improving corrections can be effectively amortized into a compact noise space transport with substantially reduced inference time computation. Figure~\ref{fig:toy_figure}(e) further visualizes the evolution of the learned correction during training. As training progresses,
the transport gradually shifts the initial noise samples toward
condition-dependent high-reward regions, increasing the average reward.

\subsection{Text-to-Image Generation}
\label{subsec:text_to_image}

\begin{figure}
    \centering
    \includegraphics[width=\linewidth]{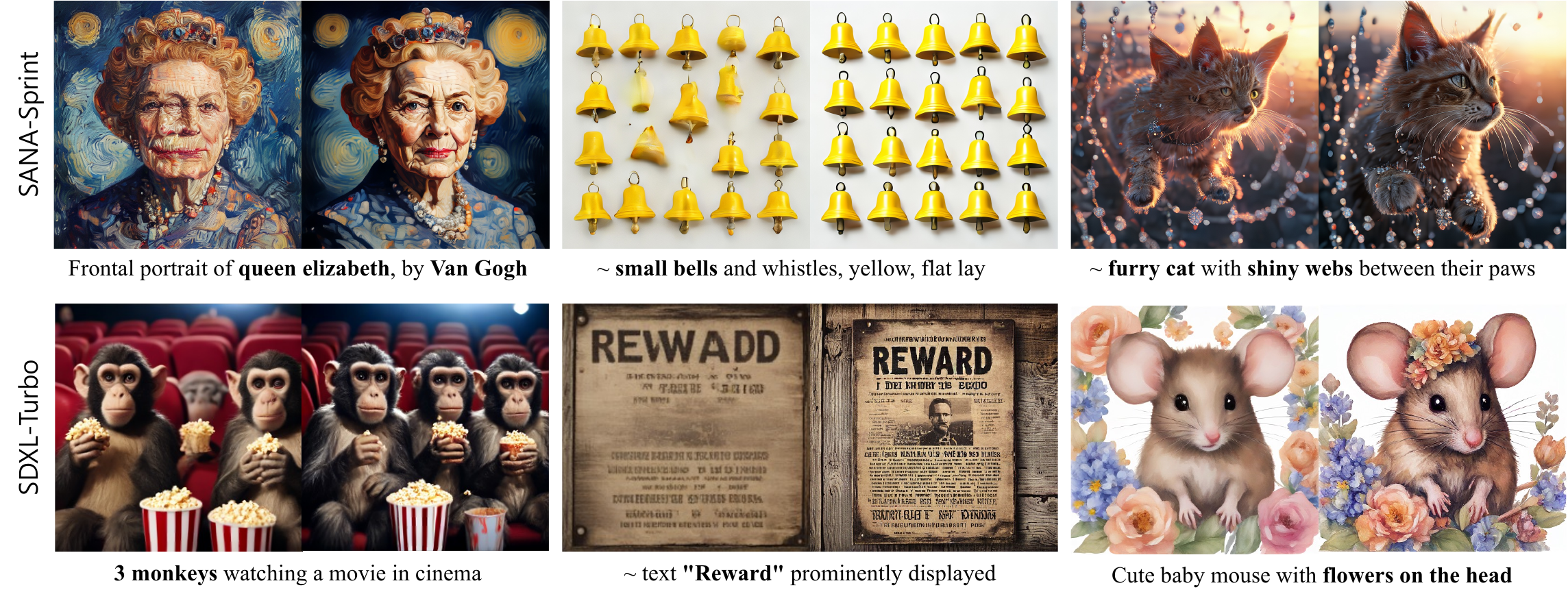}
    \vspace{-6mm}
    \caption{Qualitative comparison of one-step generations before (left) and after LeNO noise correction (right). Top: SANA-Sprint with compositional multi-reward optimization. Bottom: SDXL-Turbo with PickScore optimization.}
    \label{fig:qualitative_figure}
    \vspace{-4mm}
\end{figure}

\noindent\textbf{Experimental Setup.}
We evaluate our method on text-to-image generation under two complementary
reward-alignment settings: single reward optimization and compositional
multi-reward optimization. Across all settings, the pretrained image generator $G_\phi$ remains frozen, and only the residual noise transport is optimized. We
parameterize the residual correction $\mu_\theta$ using the same backbone
architecture as the corresponding generator, with LoRA adapters injected into
the model while keeping the underlying pretrained weights frozen. We use SDXL-Turbo~\citep{sauer2024adversarial} for the single-reward and SANA-Sprint~\citep{chen2025sana} for the compositional multi-reward setting. For single-reward optimization, we train separate residual transports for three reward functions: aesthetic score~\citep{schuhmann2022aesthetic}, PickScore~\citep{kirstain2023pickapic}, and ImageReward~\citep{xu2024imagereward}. For compositional multi-reward optimization, we optimize a composite reward consisting of ImageReward, PickScore, HPSv2~\citep{wu2023human}, and CLIPScore~\citep{hessel2021clipscore} proposed in ReNO~\citep{eyring2024reno} and use GenEval~\citep{ghosh2023geneval} solely as a held-out benchmark to evaluate compositional generalization. Full experimental details are provided in Appendix~\ref{app:implementation_details}.

\noindent\textbf{Baselines.}
We compare against both \emph{inference-time} and \emph{amortized} alignment
methods. For inference time alignment, we include Best-of-$N$, $\psi$-Sampler~\citep{yoon2026psisampler}, and ZeNO~\citep{kim2026gradient}. These methods require no additional model training, but improve reward through additional generator and reward evaluations at test time. In particular, ZeNO performs iterative gradient-free optimization independently for each initial noise, providing a direct per-sample counterpart to the optimization that our method aims to amortize. For amortized alignment, we compare with HyperNoise~\citep{eyring2026noise}, which learns to modify the initial noise distribution through differentiable reward feedback. We additionally report GoldenNoise~\citep{zhou2025golden} as a reference baseline. Unlike the target reward alignment methods considered here, GoldenNoise is trained from an offline dataset of favorable noise-sample pairs rather than directly optimizing the evaluation reward. For the held-out GenEval comparison, we further include a broad set of previously reported text-to-image models and alignment methods as reference
baselines~\citep{rombach2022high,podell2023sdxl,rafailov2023direct,batifol2025flux,esser2024scaling}.

\begin{figure}[t]
\centering

\noindent
\begin{minipage}[t]{0.46\linewidth}
\vspace{0pt}
\centering
\captionof{table}{Single reward optimization results on text-to-image generation with SDXL-Turbo.}
\vspace{-2mm}
\label{tab:single_reward}
\resizebox{\linewidth}{!}{%
\begin{tabular}{lccccc}
\toprule
Method
& $\nabla R$
& NFE $\downarrow$
& Aes. $\uparrow$
& Pick. $\uparrow$
& Img. $\uparrow$ \\
\midrule
SDXL-Turbo
& -- & 1 & 5.339 & 21.932 & 0.761 \\
\midrule
\rowcolor{gray!15}
\multicolumn{6}{l}{\textit{Inference-Time}} \\
BoN
& \xmark & 800 & 5.714 & 22.993 & 1.201 \\
$\psi$-Sampler
& \cmark & 800 & 5.945 & 23.182 & 1.421 \\
ZeNO
& \xmark & 800 & 5.750 & 23.038 & 1.260 \\
\midrule
\rowcolor{gray!15}
\multicolumn{6}{l}{\textit{Amortized}} \\
GoldenNoise\footnotemark
& -- & 1 & 5.359 & 21.908 & 0.762 \\
HyperNoise
& \cmark & 2 & 5.382 & 21.951 & 0.939 \\
\rowcolor{blue!10}
\textbf{LeNO}
& \xmark & 2 & 5.460 & 22.715 & 1.019 \\
\rowcolor{blue!10}
\textbf{LeNO + BoN}
& \xmark & 100 & \textbf{5.952} & \textbf{23.191} & \textbf{1.444} \\
\bottomrule
\end{tabular}%
}
\end{minipage}%
\hfill
\begin{minipage}[t]{0.25\linewidth}
\vspace{2mm}
\centering
\includegraphics[width=\linewidth]{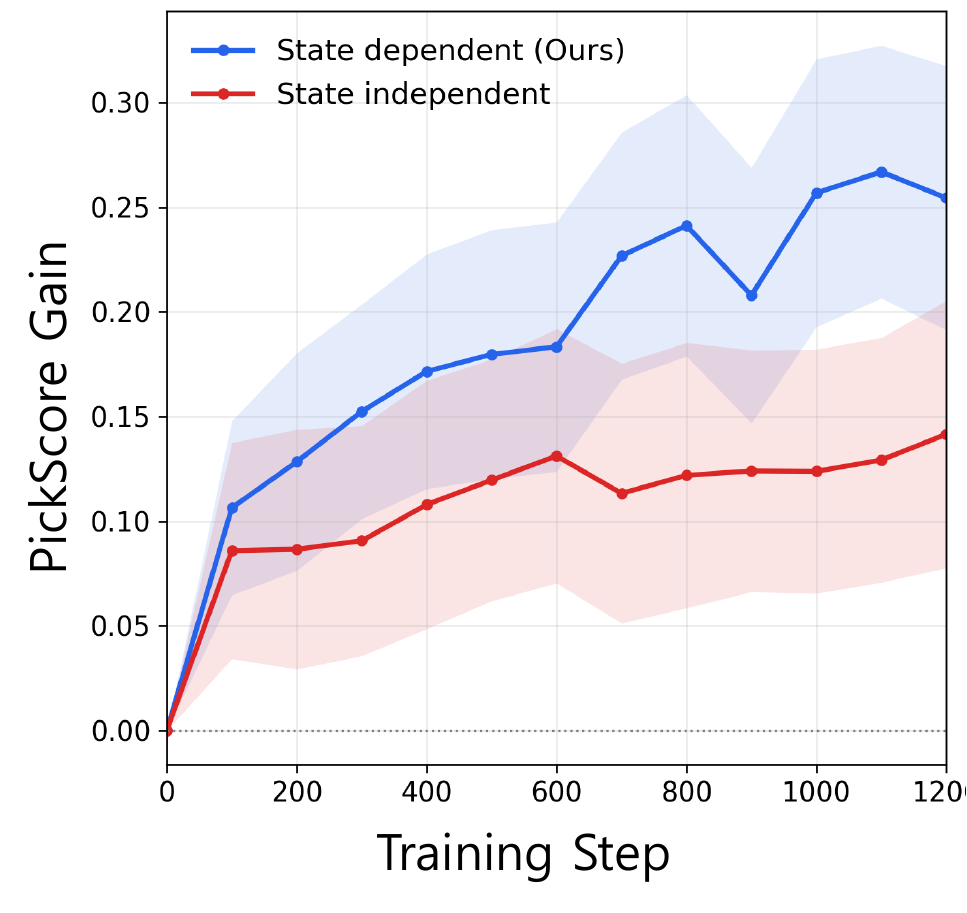}
\vspace{-5mm}
\captionof{figure}{Ablation for state-dependent training.}
\label{fig:state_dependency_ablation}
\end{minipage}
\hfill
\begin{minipage}[t]{0.25\linewidth}
\vspace{2mm}
\centering
\includegraphics[width=\linewidth]{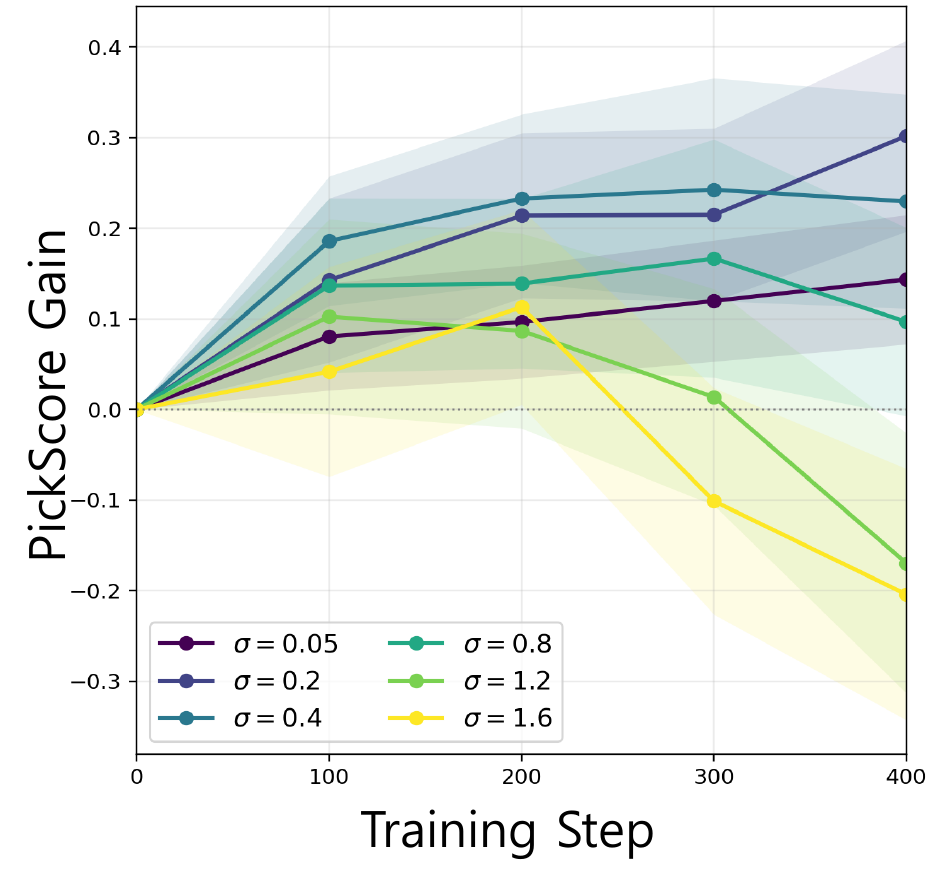}
\vspace{-5mm}
\captionof{figure}{Ablation for exploration scale $\sigma$.}
\label{fig:sigma_ablation}
\end{minipage}%
\vspace{-0.3em}

\captionof{table}{GenEval results for held-out compositional evaluation.
LeNO and HyperNoise are trained from SANA-Sprint with the same multi-reward
objective; GenEval is not used for training. HyperNoise requires reward gradients, whereas LeNO uses only scalar rewards.}
\label{tab:multi_reward}
\resizebox{\linewidth}{!}{%
\begin{tabular}{lccccccccc}
\toprule
Model
& Params (B)
& Time (s) $\downarrow$
& Mean $\uparrow$
& Single $\uparrow$
& Two $\uparrow$
& Counting $\uparrow$
& Colors $\uparrow$
& Position $\uparrow$
& Attribution $\uparrow$ \\
\midrule
SD v2.1
& 0.8 & 1.9 & 0.50 & 0.98 & 0.51 & 0.44 & 0.85 & 0.07 & 0.17 \\
SDXL
& 2.6 & 6.9 & 0.55 & 0.98 & 0.74 & 0.39 & 0.85 & 0.15 & 0.23 \\
DPO-SDXL
& 2.6 & 6.9 & 0.59 & 0.99 & 0.84 & 0.49 & 0.87 & 0.13 & 0.24 \\
Hyper-SDXL
& 2.6 & 0.3 & 0.56 & 1.00 & 0.76 & 0.43 & 0.87 & 0.10 & 0.21 \\
Flux-dev
& 12.0 & 23.0 & 0.68 & 0.99 & 0.85 & 0.74 & 0.79 & 0.21 & 0.48 \\
SD3-Medium
& 2.0 & 4.4 & 0.70 & 1.00 & 0.90 & 0.72 & 0.87 & 0.31 & 0.66 \\
\midrule
SANA-Sprint
& 0.6 & 0.2 & 0.67 & 1.00 & 0.68 & 0.60 & 0.85 & 0.40 & 0.47 \\
+ HyperNoise
& 0.9 & 0.3 & 0.71 & 1.00 & 0.75 & 0.67 & 0.84 & 0.48 & 0.53 \\
\rowcolor{blue!10}
\textbf{+ LeNO}
& 0.9 & 0.3 & 0.70 & 1.00 & 0.74 & 0.66 & 0.85 & 0.41 & 0.54 \\
\bottomrule
\end{tabular}%
}
\vspace{-3mm}
\end{figure}

\noindent\textbf{Comparison with Baselines.}
Table~\ref{tab:single_reward} reports direct optimization results for
aesthetic score, PickScore, and ImageReward. With a single residual
correction, LeNO improves all three rewards over frozen SDXL-Turbo
and achieves the best results among the evaluated amortized methods,
outperforming HyperNoise without reward gradients. GoldenNoise,
trained with different offline supervision, yields only marginal gains.
We further evaluate cross-reward transfer, held-out quality metrics, and within-prompt diversity throughout training (Appendix~\ref{app:quality_diversity}), which indicate that these gains are not obtained at the cost of perceptual quality or
sample diversity, suggesting no evident reward hacking.

Inference-time methods achieve higher absolute rewards with substantially
larger search budgets: Best-of-$N$, $\psi$-Sampler, and ZeNO use 800
generator evaluations per sample, compared with two NFEs for LeNO.
LeNO amortizes this optimization into training, avoiding repeated
reward evaluations at inference. Combining LeNO with Best-of-$N$ over
50 corrected candidates achieves competitive or better rewards than
these 800-NFE baselines using only 100 NFEs, demonstrating more
efficient inference-time scaling.

Table~\ref{tab:multi_reward} evaluates transfer to GenEval, which is
not directly optimized during multi-reward training. On SANA-Sprint,
LeNO improves the mean score from $0.67$ to $0.70$, with gains in
two-object generation, counting, position, and attribute binding.
This is comparable to HyperNoise ($0.71$), despite using only scalar
reward evaluations rather than reward gradients, and suggests
transfer beyond the training reward ensemble.

\footnotetext{GoldenNoise is trained on a pre-constructed noise--sample pair
dataset and is not directly comparable to methods that optimize the
target reward. We evaluate the SDXL-turbo model via zero-shot transfer from
SDXL.}

\noindent\textbf{State Dependency Ablation.}
We further evaluate whether conditioning the residual correction on the
current initial noise is necessary for effective amortization by comparing
the standard state-dependent model $\mu_\theta(\vz,c)$ with a
state-independent variant $\mu_\theta(c)$. To preserve the same network
architecture, the latter is implemented by feeding a fixed dummy noise
$\vz_{\mathrm{fixed}}$ in place of the current noise input, while keeping the
actual initial noise used for generation and the Local GRPO training procedure
unchanged. As shown in Fig.~\ref{fig:state_dependency_ablation}, the state-dependent
model continues to improve throughout training, whereas the state-independent
variant shows only a modest initial gain before quickly saturating. This behavior aligns with the intuition from Sec.~\ref{subsec:local_grpo}, where a shared
correction aggregates reward-improvement signals across the initial noise
population, while $\mu_\theta(\vz,c)$ can represent distinct local corrections
for different noise states.

\noindent\textbf{Exploration Scale Analysis.}
Fig.~\ref{fig:sigma_ablation} analyzes the effect of the exploration scale
$\sigma$ using SDXL-Turbo with PickScore as the training reward. We observe a
clear trade-off between exploration and locality. When $\sigma$ is too small
(e.g., $\sigma \leq 0.1$), the sampled particles remain close to the predicted
correction, yielding weak reward contrast and slower improvement. Increasing
$\sigma$ broadens the explored neighborhood and strengthens the relative reward
signal, with the best performance obtained around $\sigma=0.2$. However,
excessively large exploration degrades training; when $\sigma$ exceeds
approximately $1.2$, the reward gain collapses and can even become negative.
At this scale, the stochastic perturbation dominates the predicted correction,
weakening the local correspondence between sampled particles and the reward
landscape around the shared anchor. These results suggest that effective Local
GRPO requires sufficient exploration to reveal reward differences while
preserving a coherent local neighborhood for relative credit assignment.

\subsection{Protein Structure Generation}

\begin{figure}[t]
\centering

\begin{minipage}[t]{0.57\linewidth}
\vspace{0pt}
\centering
\captionof{table}{
Comparison of the base models and LeNO. Results are averaged over lengths 60--128, with 10 backbones per length and 8 sequences per backbone.}
\label{tab:rmf_grpo_results}

\resizebox{\linewidth}{!}{
\begin{tabular}{@{}lcccccc@{}}
\toprule
\multirow{2}{*}{Model}
& \multirow{2}{*}{NFE}
& \multirow{2}{*}{\#Query}
& \multicolumn{2}{c}{Designability}
& \multicolumn{2}{c}{Diversity} \\
\cmidrule(lr){4-5}
\cmidrule(lr){6-7}
& &
& \makecell{$<2\text{\AA}$\\$(\uparrow)$}
& \makecell{scRMSD\\$(\downarrow)$}
& \makecell{Max. Cluster\\$(\uparrow)$}
& \makecell{Pairwise scTM\\$(\downarrow)$} \\
\midrule
FrameDiff    & 5 & 0 & 0.09 & 6.19 & 0.54 & 0.24 \\
FrameFlow    & 5 & 0 & 0.04 & 6.53 & 0.68 & 0.22 \\
\midrule
RMF-M                 & 1   & 0   & 0.15 & 4.91 & 0.66 & 0.34 \\
\quad + ZeNO          & 480 & 480 & 0.28 & 3.54 & 0.65 & 0.35 \\
\rowcolor{blue!10}
\quad + \textbf{LeNO} & 2   & \textbf{0} & \textbf{0.31} & \textbf{3.27} & 0.64 & 0.38 \\
\midrule
RMF-L                 & 1   & 0   & 0.29 & 3.38 & 0.66 & 0.35 \\
\quad + ZeNO          & 480 & 480 & 0.40  & \textbf{2.60} & 0.71  & 0.36   \\
\rowcolor{blue!10}
\quad + \textbf{LeNO} & 2   & \textbf{0} & \textbf{0.42} & 2.95 & 0.66 & 0.38 \\
\bottomrule
\end{tabular}
}
\end{minipage}
\hfill
\begin{minipage}[t]{0.40\linewidth}
\vspace{0pt}
\centering
\includegraphics[width=\linewidth]{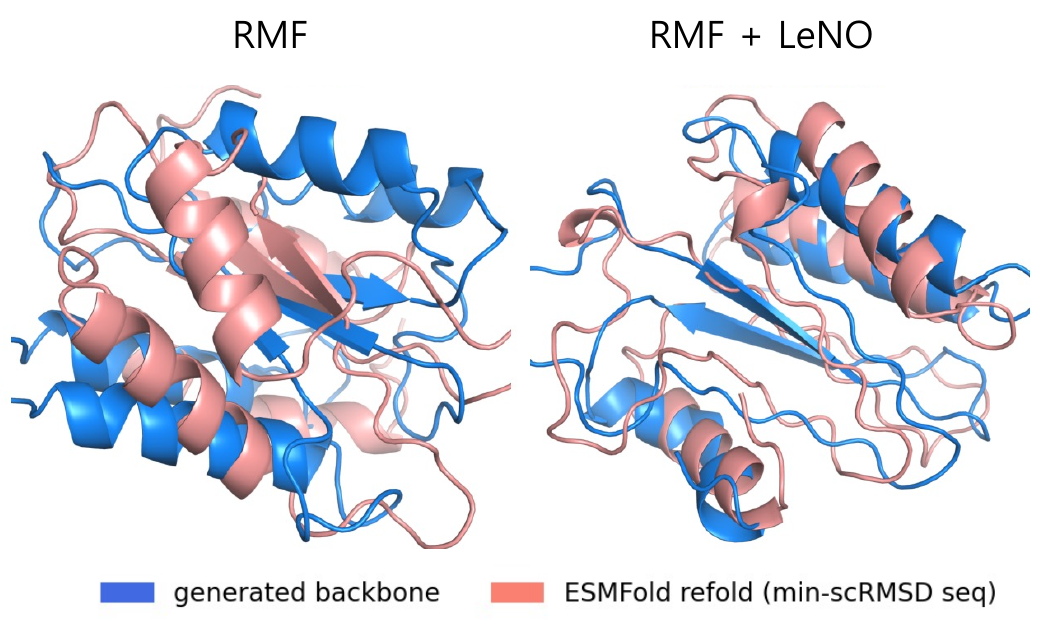}
\captionof{figure}{
Visualization of protein structures generated by the base RMF and LeNO.
}
\label{fig:protein_figure}
\end{minipage}
\vspace{-4mm}
\end{figure}

Protein backbone generation is the setting that most directly motivates LeNO.
Evaluating a backbone requires designing compatible amino-acid sequences and
predicting their folded structures through an external pipeline, so the reward
is both expensive and non-differentiable with respect to the generator.
Gradient-based amortization such as HyperNoise is therefore inapplicable, while
per-sample reward-guided search must invoke the full pipeline many times for
every test sample. LeNO instead uses this feedback only during training and
applies a single learned correction to new samples without invoking the
evaluator at inference.

We optimize \emph{designability} using the standard self-consistency pipeline.
For each generated backbone, ProteinMPNN~\citep{proteinMPNN} designs
compatible amino-acid sequences, which are then refolded using
ESMFold~\citep{ESMFold}. A backbone is considered designable when the
self-consistency RMSD (scRMSD) between the generated and refolded structures
is below $2\,\text{\AA}$. Each reward query thus involves multiple sequence
design and structure prediction calls, which dominates the cost of generation
itself.

We apply LeNO to a pretrained one-step Riemannian MeanFlow
(RMF)~\citep{woo2026riemannianmeanflow} model while keeping the generator
frozen, parameterizing the residual transport with LoRA adapters trained using
the same Local GRPO objective. As the primary baselines, we compare against
per-sample test-time ZeNO~\citep{kim2026gradient} applied to the same frozen RMF. We additionally report FrameDiff~\citep{yim2023se} and
FrameFlow~\citep{yim2023fast}, on which RMF is built, as few-step reference
models. During training, protein lengths are uniformly sampled from 60 to 128
residues, and all methods are evaluated over the same length range.

\noindent\textbf{Results.}
Table~\ref{tab:rmf_grpo_results} reports designability and diversity for
few-step protein structure generation. FrameDiff and FrameFlow degrade
substantially in the few-step regime, whereas LeNO markedly improves the
designability of both RMF variants with a single residual correction.
Notably, on RMF-M, LeNO outperforms per-sample ZeNO in both designability
and scRMSD, even though ZeNO performs 30 optimization steps with 16 particles, i.e., 480 black-box reward queries per sample, each invoking ProteinMPNN and ESMFold. LeNO instead requires no reward queries at inference, matching or exceeding explicit
per-sample search at a small fraction of its test-time cost.
Its one-time training cost (1k iterations $\times$ 36 particles $\approx$ 36k
queries) is thus amortized after generating roughly 75 backbones, beyond which
LeNO is strictly cheaper than per-sample optimization.
These gains come at a modest cost in structural diversity: MaxCluster~\citep{herbert2008maxcluster}
remains comparable to the base models, while pairwise scTM increases slightly, 
suggesting that LeNO shifts the backbone distribution toward more designable regions without collapsing it.
Qualitative examples are shown in Fig.~\ref{fig:protein_figure}, and
implementation details are provided in Appendix~\ref{app:protein_rmf_implementation}.

\section{Conclusion}

We presented LeNO as an efficient framework for learning reward-improving
corrections directly in the initial noise space of frozen deterministic
generators. Across text-to-image and protein structure generation, our results
show that a lightweight stochastic residual transport can effectively amortize costly
per-sample reward optimization while supporting non-differentiable black-box
objectives. LeNO also maintains favorable held-out behavior, including
compositional generalization in image generation and structural diversity in
protein generation. A remaining limitation is that training still requires
multiple generator and reward evaluations per anchor, which can be expensive
when the reward pipeline itself is costly. Nevertheless, the learned correction
introduces only modest inference overhead, making noise space optimization
particularly attractive when extremely fast few-step or one-step generation
is essential. 

\subsection*{AI use statement}

In this work, we used generative AI tools for sentence-level editing of texts written by authors to improve readability. We have not used generative AI tools for other aspects, such as generating synthetic data, developing theoretical models, formulating mathematical claims or proposing research hypothesis. 
We have reviewed all AI-assisted work.  We take responsibility for the final content of this work, including text, claims or artifacts produced with the aid of generative AI.

\bibliography{iclr2027_conference}

@String(ICCV= {Int. Conf. Comput. Vis.})

@String(AAAI = {AAAI})

@inproceedings{rombach2022high,
  title={High-resolution image synthesis with latent diffusion models},
  author={Rombach, Robin and Blattmann, Andreas and Lorenz, Dominik and Esser, Patrick and Ommer, Bj{\"o}rn},
  booktitle={Proceedings of the IEEE/CVF Conference on Computer Vision and Pattern Recognition},
  pages={10684--10695},
  year={2022}
}

@inproceedings{zhang2018unreasonable,
  title={The unreasonable effectiveness of deep features as a perceptual metric},
  author={Zhang, Richard and Isola, Phillip and Efros, Alexei A and Shechtman, Eli and Wang, Oliver},
  booktitle={Proceedings of the IEEE conference on computer vision and pattern recognition},
  pages={586--595},
  year={2018}
}

@article{podell2023sdxl,
  title={Sdxl: Improving latent diffusion models for high-resolution image synthesis},
  author={Podell, Dustin and English, Zion and Lacey, Kyle and Blattmann, Andreas and Dockhorn, Tim and M{\"u}ller, Jonas and Penna, Joe and Rombach, Robin},
  journal={arXiv preprint arXiv:2307.01952},
  year={2023}
}

@article{song2023consistency,
  title={Consistency models},
  author={Song, Yang and Dhariwal, Prafulla and Chen, Mark and Sutskever, Ilya},
  journal={arXiv preprint arXiv:2303.01469},
  year={2023}
}

@article{xu2024imagereward,
  title={Imagereward: Learning and evaluating human preferences for text-to-image generation},
  author={Xu, Jiazheng and Liu, Xiao and Wu, Yuchen and Tong, Yuxuan and Li, Qinkai and Ding, Ming and Tang, Jie and Dong, Yuxiao},
  journal={Advances in Neural Information Processing Systems},
  volume={36},
  year={2024}
}

@inproceedings{esser2024scaling,
  title={Scaling rectified flow transformers for high-resolution image synthesis},
  author={Esser, Patrick and Kulal, Sumith and Blattmann, Andreas and Entezari, Rahim and M{\"u}ller, Jonas and Saini, Harry and Levi, Yam and Lorenz, Dominik and Sauer, Axel and Boesel, Frederic and others},
  booktitle={Forty-first international conference on machine learning},
  year={2024}
}

@article{batifol2025flux,
  title={FLUX. 1 Kontext: Flow Matching for In-Context Image Generation and Editing in Latent Space},
  author={Batifol, Stephen and Blattmann, Andreas and Boesel, Frederic and Consul, Saksham and Diagne, Cyril and Dockhorn, Tim and English, Jack and English, Zion and Esser, Patrick and Kulal, Sumith and others},
  journal={arXiv e-prints},
  pages={arXiv--2506},
  year={2025}
}

@inproceedings{
yoon2026psisampler,
title={\${\textbackslash}Psi\$-Sampler: Initial Particle Sampling for {SMC}-Based Inference-Time Reward Alignment in Score Models},
author={TaeHoon Yoon and Yunhong Min and Kyeongmin Yeo and Minhyuk Sung},
booktitle={The Thirty-ninth Annual Conference on Neural Information Processing Systems},
year={2026},
url={https://openreview.net/forum?id=slVqJAI5sT}
}

@inproceedings{sauer2024adversarial,
  title={Adversarial diffusion distillation},
  author={Sauer, Axel and Lorenz, Dominik and Blattmann, Andreas and Rombach, Robin},
  booktitle={European Conference on Computer Vision},
  pages={87--103},
  year={2024},
  organization={Springer}
}

@inproceedings{
liu2026flowgrpo,
title={Flow-{GRPO}: Training Flow Matching Models via Online {RL}},
author={Jie Liu and Gongye Liu and Jiajun Liang and Yangguang Li and Jiaheng Liu and Xintao Wang and Pengfei Wan and Di ZHANG and Wanli Ouyang},
booktitle={The Thirty-ninth Annual Conference on Neural Information Processing Systems},
year={2026},
url={https://openreview.net/forum?id=oCBKGw5HNf}
}

@article{lee2025pcpo,
  title={PCPO: Proportionate Credit Policy Optimization for Aligning Image Generation Models},
  author={Lee, Jeongjae and Ye, Jong Chul},
  journal={arXiv preprint arXiv:2509.25774},
  year={2025}
}

@inproceedings{
eyring2024reno,
title={Re{NO}: Enhancing One-step Text-to-Image Models through Reward-based Noise Optimization},
author={Luca Eyring and Shyamgopal Karthik and Karsten Roth and Alexey Dosovitskiy and Zeynep Akata},
booktitle={The Thirty-eighth Annual Conference on Neural Information Processing Systems},
year={2024},
url={https://openreview.net/forum?id=MXY0qsGgeO}
}

@inproceedings{tao2022df,
  title={DF-GAN: A Simple and Effective Baseline for Text-to-Image Synthesis},
  author={Tao, Ming and Tang, Hao and Wu, Fei and Jing, Xiao-Yuan and Bao, Bing-Kun and Xu, Changsheng},
  booktitle={Proceedings of the IEEE/CVF Conference on Computer Vision and Pattern Recognition},
  pages={16515--16525},
  year={2022}
}

@inproceedings{
ghosh2023geneval,
title={GenEval: An object-focused framework for evaluating text-to-image alignment},
author={Dhruba Ghosh and Hannaneh Hajishirzi and Ludwig Schmidt},
booktitle={Thirty-seventh Conference on Neural Information Processing Systems Datasets and Benchmarks Track},
year={2023},
url={https://openreview.net/forum?id=Wbr51vK331}
}

@inproceedings{
kirstain2023pickapic,
title={Pick-a-Pic: An Open Dataset of User Preferences for Text-to-Image Generation},
author={Yuval Kirstain and Adam Polyak and Uriel Singer and Shahbuland Matiana and Joe Penna and Omer Levy},
booktitle={Thirty-seventh Conference on Neural Information Processing Systems},
year={2023},
url={https://openreview.net/forum?id=G5RwHpBUv0}
}

@misc{schuhmann2022aesthetic,
author       = {Christoph Schuhmann},
title        = {Improved Aesthetic Predictor},
year         = {2022},
howpublished = {\url{https://github.com/christophschuhmann/improved-aesthetic-predictor}},
note         = {GitHub repository}
}

@misc{woo2026riemannianmeanflow,
title={Riemannian MeanFlow}, 
author={Dongyeop Woo and Marta Skreta and Seonghyun Park and Kirill Neklyudov and Sungsoo Ahn},
year={2026},
eprint={2602.07744},
archivePrefix={arXiv},
primaryClass={cs.LG},
url={https://arxiv.org/abs/2602.07744}, 
}

@article{
proteinMPNN,
author = {J. Dauparas  and I. Anishchenko  and N. Bennett  and H. Bai  and R. J. Ragotte  and L. F. Milles  and B. I. M. Wicky  and A. Courbet  and R. J. de Haas  and N. Bethel  and P. J. Y. Leung  and T. F. Huddy  and S. Pellock  and D. Tischer  and F. Chan  and B. Koepnick  and H. Nguyen  and A. Kang  and B. Sankaran  and A. K. Bera  and N. P. King  and D. Baker },
title = {Robust deep learning–based protein sequence design using ProteinMPNN},
journal = {Science},
volume = {378},
number = {6615},
pages = {49-56},
year = {2022},
doi = {10.1126/science.add2187},
URL = {https://www.science.org/doi/abs/10.1126/science.add2187},
eprint = {https://www.science.org/doi/pdf/10.1126/science.add2187}
}

@article{
ESMFold,
author = {Zeming Lin  and Halil Akin  and Roshan Rao  and Brian Hie  and Zhongkai Zhu  and Wenting Lu  and Nikita Smetanin  and Robert Verkuil  and Ori Kabeli  and Yaniv Shmueli  and Allan dos Santos Costa  and Maryam Fazel-Zarandi  and Tom Sercu  and Salvatore Candido  and Alexander Rives },
title = {Evolutionary-scale prediction of atomic-level protein structure with a language model},
journal = {Science},
volume = {379},
number = {6637},
pages = {1123-1130},
year = {2023},
doi = {10.1126/science.ade2574},
URL = {https://www.science.org/doi/abs/10.1126/science.ade2574},
eprint = {https://www.science.org/doi/pdf/10.1126/science.ade2574}
}

@inproceedings{yin2024one,
  title={One-step diffusion with distribution matching distillation},
  author={Yin, Tianwei and Gharbi, Micha{\"e}l and Zhang, Richard and Shechtman, Eli and Durand, Fredo and Freeman, William T and Park, Taesung},
  booktitle={Proceedings of the IEEE/CVF conference on computer vision and pattern recognition},
  pages={6613--6623},
  year={2024}
}

@article{geng2025mean,
  title={Mean flows for one-step generative modeling},
  author={Geng, Zhengyang and Deng, Mingyang and Bai, Xingjian and Kolter, J Zico and He, Kaiming},
  journal={arXiv preprint arXiv:2505.13447},
  year={2025}
}

@inproceedings{black2024training,
  title={Training diffusion models with reinforcement learning},
  author={Black, Kevin and Janner, Michael and Du, Yilun and Kostrikov, Ilya and Levine, Sergey},
  booktitle={International Conference on Learning Representations},
  volume={2024},
  pages={4965--4987},
  year={2024}
}

@article{JMLR:v15:wierstra14a,
  author  = {Daan Wierstra and Tom Schaul and Tobias Glasmachers and Yi Sun and Jan Peters and J{\"u}rgen Schmidhuber},
  title   = {Natural Evolution Strategies},
  journal = {Journal of Machine Learning Research},
  year    = {2014},
  volume  = {15},
  number  = {27},
  pages   = {949--980},
  url     = {http://jmlr.org/papers/v15/wierstra14a.html}
}

@article{kim2026gradient,
  title={Gradient-Free Noise Optimization for Reward Alignment in Generative Models},
  author={Kim, Jeongsol and Kim, Hongeun and Wang, Jian and Ye, Jong Chul},
  journal={arXiv preprint arXiv:2605.11347},
  year={2026}
}

@inproceedings{zhou2025golden,
  title={Golden noise for diffusion models: A learning framework},
  author={Zhou, Zikai and Shao, Shitong and Bai, Lichen and Zhang, Shufei and Xu, Zhiqiang and Han, Bo and Xie, Zeke},
  booktitle={2025 IEEE/CVF International Conference on Computer Vision (ICCV)},
  pages={17688--17697},
  year={2025},
  organization={IEEE}
}

@article{eyring2026noise,
  title={Noise hypernetworks: Amortizing test-time compute in diffusion models},
  author={Eyring, Luca and Karthik, Shyamgopal and Dosovitskiy, Alexey and Ruiz, Nataniel and Akata, Zeynep},
  journal={Advances in Neural Information Processing Systems},
  volume={38},
  pages={95429--95462},
  year={2026}
}

@inproceedings{he2026neighbor,
  title={Neighbor grpo: Contrastive ode policy optimization aligns flow models},
  author={He, Dailan and Feng, Guanlin and Ge, Xingtong and Niu, Yazhe and Zhang, Yi and Ma, Bingqi and Song, Guanglu and Liu, Yu and Li, Hongsheng},
  booktitle={Proceedings of the IEEE/CVF Conference on Computer Vision and Pattern Recognition},
  pages={6033--6042},
  year={2026}
}

@inproceedings{hessel2021clipscore,
  title={Clipscore: A reference-free evaluation metric for image captioning},
  author={Hessel, Jack and Holtzman, Ari and Forbes, Maxwell and Le Bras, Ronan and Choi, Yejin},
  booktitle={Proceedings of the 2021 conference on empirical methods in natural language processing},
  pages={7514--7528},
  year={2021}
}

@inproceedings{chen2025sana,
  title={Sana-sprint: One-step diffusion with continuous-time consistency distillation},
  author={Chen, Junsong and Xue, Shuchen and Zhao, Yuyang and Yu, Jincheng and Paul, Sayak and Chen, Junyu and Cai, Han and Han, Song and Xie, Enze},
  booktitle={2025 IEEE/CVF International Conference on Computer Vision (ICCV)},
  pages={16185--16195},
  year={2025},
  organization={IEEE}
}

@article{wu2023human,
  title={Human Preference Score v2: A Solid Benchmark for Evaluating Human Preferences of Text-to-Image Synthesis},
  author={Wu, Xiaoshi and Hao, Yiming and Sun, Keqiang and Chen, Yixiong and Zhu, Feng and Zhao, Rui and Li, Hongsheng},
  journal={arXiv preprint arXiv:2306.09341},
  year={2023}
}

@article{rafailov2023direct,
  title={Direct preference optimization: Your language model is secretly a reward model},
  author={Rafailov, Rafael and Sharma, Archit and Mitchell, Eric and Manning, Christopher D and Ermon, Stefano and Finn, Chelsea},
  journal={Advances in neural information processing systems},
  volume={36},
  pages={53728--53741},
  year={2023}
}

@article{yim2023se,
  title={SE (3) diffusion model with application to protein backbone generation},
  author={Yim, Jason and Trippe, Brian L and De Bortoli, Valentin and Mathieu, Emile and Doucet, Arnaud and Barzilay, Regina and Jaakkola, Tommi},
  journal={arXiv preprint arXiv:2302.02277},
  year={2023}
}

@article{yim2023fast,
  title={Fast protein backbone generation with se (3) flow matching},
  author={Yim, Jason and Campbell, Andrew and Foong, Andrew YK and Gastegger, Michael and Jim{\'e}nez-Luna, Jos{\'e} and Lewis, Sarah and Satorras, Victor Garcia and Veeling, Bastiaan S and Barzilay, Regina and Jaakkola, Tommi and others},
  journal={arXiv preprint arXiv:2310.05297},
  year={2023}
}

@misc{herbert2008maxcluster,
  author       = {Herbert, A. and Sternberg, M.},
  title        = {MaxCluster: a tool for protein structure comparison and clustering},
  year         = {2008},
  howpublished = {\url{https://ic.ac.uk}}
}

@article{oquab2023dinov2,
  title={Dinov2: Learning robust visual features without supervision},
  author={Oquab, Maxime and Darcet, Timoth{\'e}e and Moutakanni, Th{\'e}o and Vo, Huy and Szafraniec, Marc and Khalidov, Vasil and Fernandez, Pierre and Haziza, Daniel and Massa, Francisco and El-Nouby, Alaaeldin and others},
  journal={arXiv preprint arXiv:2304.07193},
  year={2023}
}

@inproceedings{ke2021musiq,
  title={Musiq: Multi-scale image quality transformer},
  author={Ke, Junjie and Wang, Qifei and Wang, Yilin and Milanfar, Peyman and Yang, Feng},
  booktitle={2021 IEEE/CVF International Conference on Computer Vision (ICCV)},
  pages={5128--5137},
  year={2021},
  organization={IEEE}
}

@inproceedings{wang2023exploring,
  title={Exploring clip for assessing the look and feel of images},
  author={Wang, Jianyi and Chan, Kelvin CK and Loy, Chen Change},
  booktitle={Proceedings of the AAAI conference on artificial intelligence},
  volume={37},
  number={2},
  pages={2555--2563},
  year={2023}
}
\bibliographystyle{iclr2027_conference}

\newpage
\appendix

\section{Relation to Inference-Time Zeroth-Order Estimator}
\label{app:zeroth_order_derivation}

We analyze the gradient of the expected reward under the
stochastic residual transport, omitting the residual-magnitude
regularizer and considering the additive transition before
norm projection.
Fix an anchor $\vz$ and condition $c$, and define the
mean-shifted anchor
\begin{equation}
\tilde{\vz}
:=
\vz+\eta\mu_\theta(\vz,c).
\end{equation}
The expected reward under Gaussian action sampling is
\begin{equation}
J_R(\vz;c)
:=
\mathbb{E}_{\epsilon\sim\mathcal{N}(0,I)}
\left[
\mathcal{R}(\tilde{\vz}+\sigma\epsilon;c)
\right]
=
\mathcal{R}_\sigma(\tilde{\vz};c).
\label{eq:reward_objective}
\end{equation}
We assume the integrability conditions required for
differentiating the Gaussian smoothing hold.
For brevity, $\nabla_{\mu_\theta}$ denotes differentiation
with respect to the predicted residual mean
$\mu_\theta(\vz,c)$, rather than the network parameters.

By the chain rule and the Gaussian score identity,
\begin{equation}
\begin{aligned}
\nabla_{\mu_\theta}J_R(\vz;c)
&=
\eta\,\nabla_{\tilde{\vz}}
\mathcal{R}_\sigma(\tilde{\vz};c)
\\
&=
\frac{\eta}{\sigma}
\mathbb{E}_{\epsilon}
\left[
\mathcal{R}(\tilde{\vz}+\sigma\epsilon;c)\epsilon
\right].
\end{aligned}
\label{eq:pg_score_estimator}
\end{equation}
Thus, the reward gradient with respect to the residual mean
equals $\eta$ times the zeroth-order gradient of the smoothed
reward at the mean-shifted anchor.

Since $\mathbb{E}_{\epsilon}[\epsilon]=0$, a baseline
independent of the sampled perturbation does not change
the expected gradient.
For $K>1$ independent perturbations, define
\begin{equation}
R_i
=
\mathcal{R}(\tilde{\vz}+\sigma\epsilon_i;c),
\qquad
\bar R
=
\frac{1}{K}\sum_{i=1}^{K}R_i.
\end{equation}
An unbiased centered estimator is
\begin{equation}
\widehat{\nabla_{\mu_\theta}J_R}
=
\frac{\eta}{(K-1)\sigma}
\sum_{i=1}^{K}(R_i-\bar R)\epsilon_i.
\label{eq:app_centered_zo}
\end{equation}
The factor $1/(K-1)$ corrects for the dependence introduced
by including $R_i$ in $\bar R$.
Equivalently, this estimator uses the leave-one-out baseline
$b_{-i}=(K-1)^{-1}\sum_{j\ne i}R_j$, since
\begin{equation}
R_i-b_{-i}
=
\frac{K}{K-1}(R_i-\bar R).
\end{equation}
Equation~\ref{eq:app_centered_zo} has the same form as
\eqref{eq:zo_estimator}, evaluated at $\tilde{\vz}$
and scaled by $\eta$.

Local GRPO instead uses standardized group-relative advantages,
\begin{equation}
A_i
=
\frac{R_i-\bar R}{s_R+\varepsilon}.
\end{equation}
At the sampling policy $\theta=\theta_{\mathrm{old}}$,
the unclipped reward-gradient estimate with respect to
the residual mean is
\begin{equation}
\frac{\eta}{K\sigma}\sum_{i=1}^{K}A_i\epsilon_i
=
\frac{K-1}{K(s_R+\varepsilon)}
\widehat{\nabla_{\mu_\theta}J_R}.
\end{equation}
For a fixed sampled group, the multiplier is positive,
so normalization preserves the direction of this local
estimate.
Across groups, however, normalization changes their relative
weights and does not generally preserve the population
reward gradient.
Subsequent policy updates introduce likelihood-ratio
weighting and potentially PPO clipping, so the resulting
GRPO updates need not remain proportional to the
zeroth-order estimate.

\section{Proof of Proposition~\ref{prop:distributional_kl}}
\label{app:distributional_kl}

\distributionalkl*

\begin{proof}
Fix a condition $c$ and $\sigma>0$.
Let $p_0=\mathcal{N}(0,I_d)$ and
\[
\mathcal{T}(\vz,\va)
=
\|\vz\|_2
\frac{\vz+\va}{\|\vz+\va\|_2}.
\]
We first show that zero-mean Gaussian actions preserve
the prior, then apply data processing to establish
the KL bound.

\paragraph{Preservation of the Gaussian prior.}
Let $\vz,\epsilon\sim\mathcal{N}(0,I_d)$ be independent.
The output under zero-mean residual actions is
\begin{equation}
\vy
=
\mathcal{T}(\vz,\sigma\epsilon)
=
\|\vz\|_2
\frac{\vz+\sigma\epsilon}
{\|\vz+\sigma\epsilon\|_2}.
\label{eq:app_reference_noise}
\end{equation}
The denominator is nonzero almost surely, and the
transition may be defined arbitrarily on the null set
where it vanishes.

Write $\vz=r\vu$, where $r=\|\vz\|_2$.
For an isotropic Gaussian, $\vu$ is uniform on the unit
sphere and independent of $r$.
Conditional on $r>0$, the vector $r\vu+\sigma\epsilon$
is rotationally invariant: for every orthogonal matrix
$Q$, the pairs $(Q\vu,Q\epsilon)$ and $(\vu,\epsilon)$
have the same conditional distribution.
Its normalized direction is therefore uniform on the
unit sphere conditional on $r$.
Since this conditional distribution does not depend on
$r$, the output direction is independent of the radius.

The transition also preserves the radius exactly,
$\|\vy\|_2=r$.
Thus, $\vy$ has the same radial distribution and
independent uniform direction as $\vz$, giving
\begin{equation}
\mathcal{T}(\vz,\sigma\epsilon)
\sim \mathcal{N}(0,I_d)=p_0.
\label{eq:app_reference_prior}
\end{equation}

\paragraph{Bounding the marginal KL divergence.}
Define the joint distributions of anchors and residual
actions under the learned and zero-mean transports by
\begin{equation}
\begin{aligned}
P_\theta(d\vz,d\va\mid c)
&=
p_0(d\vz)\,
\mathcal{N}\!\left(
d\va;\eta\mu_\theta(\vz,c),\sigma^2 I
\right),
\\
P_{\mathrm{ref}}(d\vz,d\va)
&=
p_0(d\vz)\,
\mathcal{N}\!\left(
d\va;0,\sigma^2 I
\right).
\end{aligned}
\label{eq:app_joint_exploration}
\end{equation}
Both joint distributions share the anchor marginal $p_0$.
The chain rule for KL divergence therefore gives
\begin{equation}
\begin{aligned}
D_{\mathrm{KL}}\!\left(
P_\theta(\cdot\mid c)\Vert P_{\mathrm{ref}}
\right)
&=
\mathbb{E}_{\vz\sim p_0}
D_{\mathrm{KL}}\!\left(
\mathcal{N}(\eta\mu_\theta(\vz,c),\sigma^2 I)
\,\Vert\,
\mathcal{N}(0,\sigma^2 I)
\right)
\\
&=
\frac{\eta^2}{2\sigma^2}
\mathbb{E}_{\vz\sim p_0}
\|\mu_\theta(\vz,c)\|_2^2.
\end{aligned}
\label{eq:app_joint_kl}
\end{equation}
The second equality follows from the Gaussian KL formula
for identical covariances.
The assumed finite second moment makes this divergence
finite.

Applying the same map
$(\vz,\va)\mapsto\mathcal{T}(\vz,\va)$ to
$P_\theta(\cdot\mid c)$ produces $q_\theta(\cdot\mid c)$.
By \eqref{eq:app_reference_prior}, applying it to
$P_{\mathrm{ref}}$ produces $p_0$.
The data-processing inequality consequently yields
\begin{equation}
\begin{aligned}
D_{\mathrm{KL}}\!\left(
q_\theta(\cdot\mid c)\Vert p_0
\right)
&\leq
D_{\mathrm{KL}}\!\left(
P_\theta(\cdot\mid c)\Vert P_{\mathrm{ref}}
\right)
\\
&=
\frac{\eta^2}{2\sigma^2}
\mathbb{E}_{\vz\sim p_0}
\|\mu_\theta(\vz,c)\|_2^2.
\end{aligned}
\label{eq:app_marginal_kl_bound}
\end{equation}
This proves the proposition without requiring an
invertible transition or evaluating its Jacobian.
\end{proof}

\paragraph{From the reward-tilted target to the auxiliary objective.}
Let $\lambda>0$ and assume
$0<Z(c)<\infty$ and
$\mathbb{E}_{\vz'\sim q_\theta(\cdot\mid c)}
[|\mathcal{R}(\vz';c)|]<\infty$.
From the definition of $p^\star$,
\begin{equation}
\log\frac{q_\theta(\vz'\mid c)}{p^\star(\vz'\mid c)}
=
\log\frac{q_\theta(\vz'\mid c)}{p_0(\vz')}
-
\frac{\mathcal{R}(\vz';c)}{\lambda}
+
\log Z(c).
\end{equation}
Taking expectations under $q_\theta(\cdot\mid c)$ gives
\begin{equation}
\begin{aligned}
D_{\mathrm{KL}}\!\left(
q_\theta(\cdot\mid c)\Vert p^\star(\cdot\mid c)
\right)
&=
D_{\mathrm{KL}}\!\left(
q_\theta(\cdot\mid c)\Vert p_0
\right)
\\
&\quad-
\frac{1}{\lambda}
\mathbb{E}_{\vz'\sim q_\theta(\cdot\mid c)}
[\mathcal{R}(\vz';c)]
+
\log Z(c).
\end{aligned}
\label{eq:app_target_kl_decomposition}
\end{equation}
Since $Z(c)$ is independent of $\theta$, minimizing
this divergence is equivalent to maximizing
$\mathcal{L}(q_\theta;c)$.

Applying Proposition~\ref{prop:distributional_kl} to
the prior-relative KL term yields
\begin{equation}
\begin{aligned}
\mathcal{L}(q_\theta;c)
&=
\mathbb{E}_{\vz'\sim q_\theta(\cdot\mid c)}
[\mathcal{R}(\vz';c)]
-
\lambda D_{\mathrm{KL}}\!\left(
q_\theta(\cdot\mid c)\Vert p_0
\right)
\\
&\geq
\mathbb{E}_{\vz'\sim q_\theta(\cdot\mid c)}
[\mathcal{R}(\vz';c)]
-
\frac{\lambda\eta^2}{2\sigma^2}
\mathbb{E}_{\vz\sim p_0}
\|\mu_\theta(\vz,c)\|_2^2
\\
&=
\mathcal{L}_{\mathrm{aux}}(\theta;c).
\end{aligned}
\label{eq:app_distributional_lower_bound}
\end{equation}
This establishes
\eqref{eq:distributional_lower_bound}.
The same stochastic transport and sampling scale are
used at inference, so the bound applies to the corrected
noise distribution used for generation.

\section{State-Dependent versus Shared Corrections}
\label{app:state_dependency_justification}

\paragraph{First-order surrogate.}
Fix a condition $c$, consider the additive transition before projection, and
set $\eta=1$ for notational simplicity. We write $\mu$ for an unrestricted
correction function, independently of its neural parameterization. Under this
transition, the corrected noise is $\vz'=\vz+\mu(\vz,c)+\sigma\epsilon$, and
the expected reward over the Gaussian action equals the smoothed reward at the
mean-shifted anchor (\eqref{eq:reward_objective}). The auxiliary objective in
\eqref{eq:distributional_lower_bound} then reads
\begin{equation}
\mathcal{L}_{\mathrm{aux}}(\mu;c)
=
\mathbb{E}_{\vz\sim p_0}\!\left[
\mathcal{R}_\sigma\big(\vz+\mu(\vz,c);c\big)
-
\frac{\lambda}{2\sigma^2}\|\mu(\vz,c)\|_2^2
\right].
\end{equation}
Linearizing $\mathcal{R}_\sigma$ around the anchor,
$\mathcal{R}_\sigma(\vz+\mu;c)\approx\mathcal{R}_\sigma(\vz;c)
+g(\vz,c)^\top\mu(\vz,c)$ with
$g(\vz,c)=\nabla_{\vz}\mathcal{R}_\sigma(\vz;c)$ as in
\eqref{eq:zo_estimator}, and dropping the $\mu$-independent term gives
\begin{equation}
J_{\mathrm{lin}}(\mu;c)
:=
\mathbb{E}_{\vz\sim p_0}
\left[
g(\vz,c)^\top\mu(\vz,c)
-
\frac{\alpha}{2}\|\mu(\vz,c)\|_2^2
\right],
\qquad \alpha=\frac{\lambda}{\sigma^2}.
\label{eq:state_dependent_objective}
\end{equation}
The following statements are exact for this surrogate, rather than for the
original nonlinear reward objective.

\begin{lemma}
\label{lemma:state_dependent}
Assume $\mathbb{E}_{p_0}\|g(\vz,c)\|_2^2<\infty$.
Let $J_{\mathrm{state}}^\star$ and $J_{\mathrm{shared}}^\star$ denote the
optimal values of \eqref{eq:state_dependent_objective} over square-integrable
state-dependent corrections $\mu(\vz,c)$ and shared corrections $\mu(c)$,
respectively. The optimal corrections are
$\mu_{\mathrm{state}}^\star(\vz,c)=g(\vz,c)/\alpha$ and
$\mu_{\mathrm{shared}}^\star(c)=\mathbb{E}_{p_0}[g(\vz,c)]/\alpha$, and
\begin{equation}
\Delta J
:=
J_{\mathrm{state}}^\star-J_{\mathrm{shared}}^\star
=
\frac{1}{2\alpha}
\operatorname{Tr}\!\left(
\operatorname{Cov}_{\vz\sim p_0}[g(\vz,c)]
\right)
\geq 0,
\label{eq:state_dependent_gap}
\end{equation}
with equality if and only if $g(\vz,c)$ is constant $p_0$-almost surely.
\end{lemma}

\begin{proof}
For each $\vz$, completing the square gives
\begin{equation}
g^\top\mu-\frac{\alpha}{2}\|\mu\|_2^2
=
\frac{1}{2\alpha}\|g\|_2^2
-\frac{\alpha}{2}\left\|\mu-\frac{g}{\alpha}\right\|_2^2,
\end{equation}
where we omit the arguments $(\vz,c)$. The pointwise maximizer is
$\mu_{\mathrm{state}}^\star=g/\alpha$, which is square-integrable by the
finite second-moment assumption, and
$J_{\mathrm{state}}^\star=\frac{1}{2\alpha}\mathbb{E}_{p_0}\|g\|_2^2$.
For a shared correction $\mu(c)$, the objective reduces to
$\mathbb{E}_{p_0}[g]^\top\mu(c)-\frac{\alpha}{2}\|\mu(c)\|_2^2$, and the same
argument gives $\mu_{\mathrm{shared}}^\star=\mathbb{E}_{p_0}[g]/\alpha$ and
$J_{\mathrm{shared}}^\star=\frac{1}{2\alpha}\|\mathbb{E}_{p_0}[g]\|_2^2$.
Subtracting,
\begin{equation}
\Delta J
=
\frac{1}{2\alpha}
\left(
\mathbb{E}_{p_0}\|g\|_2^2-\|\mathbb{E}_{p_0}[g]\|_2^2
\right)
=
\frac{1}{2\alpha}
\operatorname{Tr}\!\left(\operatorname{Cov}_{p_0}[g]\right)
\geq 0,
\end{equation}
with equality if and only if $\operatorname{Cov}_{p_0}[g]=0$, i.e., $g$ is
constant $p_0$-almost surely.
\end{proof}

\newpage
\section{Implementation Details}
\label{app:implementation_details}

All experiments are conducted using NVIDIA RTX 4090 GPUs with 24GB of memory.

\subsection{Text-to-Image Generation}

\paragraph{Hyperparameters.}
We use the same hyperparameters for all text-to-image Local GRPO experiments. We set the correction scale to $\eta=0.3$ and the exploration standard deviation to $\sigma=0.2$, using 32 particles per group, 16 prompt anchors per update, and four GRPO epochs per rollout. Advantages are normalized using the group mean and standard deviation and clipped to $[-5,5]$. The PPO probability ratio is clipped to $[0.8,1.2]$, while the log-ratio is clipped to $[-10,10]$ for numerical stability.
\vspace{-3mm}
\begin{table}[H]
\centering
\caption{GRPO training hyperparameters.}
\label{tab:grpo_hyperparameters}
\begin{tabular}{lc}
\toprule
\textbf{Hyperparameter} & \textbf{Value} \\
\midrule
Correction scale $\eta$          & $0.3$ \\
Exploration std.\ $\sigma$       & $0.2$ \\
KL regularization weight $\beta$  & $0.01$ \\
Global particle group size       & $32$ \\
Prompt anchors per update        & $16$ \\
GRPO epochs per rollout          & $4$ \\
Advantage normalization          & Group mean/std \\
Advantage clipping               & $[-5,\,5]$ \\
PPO ratio clipping               & $[0.8,\,1.2]$ ($\epsilon=0.2$) \\
Log-ratio clipping               & $[-10,\,10]$ \\
\bottomrule
\end{tabular}
\end{table}

\paragraph{Single Reward Optimization.}
For single-reward optimization, we use SDXL-Turbo as the frozen generator and
consider three reward functions: the aesthetic score predictor, PickScore,
and ImageReward. We train a separate LeNO model for each reward using only the
corresponding scalar reward signal. For a fair comparison, we also retrain
HyperNoise independently for all three reward functions rather than reusing a
single pretrained checkpoint. In each case, we follow the default HyperNoise
training configuration and change only the target reward function. The Local
GRPO hyperparameters and model-specific training configurations are summarized
in Tab.~\ref{tab:grpo_hyperparameters} and Tab.~\ref{tab:training_config}, respectively.

\paragraph{Compositional Multi-Reward Optimization.}
For compositional optimization, we follow the multi-reward formulation of
ReNO~\citep{eyring2024reno} and use SANA-Sprint as the frozen generator. Specifically,
the training reward combines ImageReward, HPSv2, PickScore, and CLIPScore,
which provide complementary signals for human preference, perceptual quality,
and text--image consistency. Following ReNO, we use the weighted objective
\begin{equation}
R_{\mathrm{multi}}
=
1.0\,R_{\mathrm{ImageReward}}
+
5.0\,R_{\mathrm{HPSv2}}
+
0.05\,R_{\mathrm{PickScore}}
+
1.0\,R_{\mathrm{CLIPScore}},
\end{equation}
where the weighting accounts for the substantially different numerical scales
of the individual reward models. The corresponding Local GRPO and training
configurations are reported in
Tab.~\ref{tab:grpo_hyperparameters} and Tab.~\ref{tab:training_config}.

To evaluate whether improvements transfer beyond the reward models used for
training, we additionally report GenEval~\citep{ghosh2023geneval} as a held-out
evaluation. The GenEval prompts and evaluator are not used during LeNO
training, and the GenEval score is not included in the multi-reward objective.
We evaluate the trained model only after optimization using the standard
GenEval benchmark, which measures compositional text--image alignment across
single-object generation, two-object composition, counting, color, spatial
relations, and attribute binding. Thus, improvements on GenEval indicate that
the learned noise correction generalizes beyond direct optimization of the
four training rewards.

\begin{table}[H]
\centering
\caption{Training configurations for HyperNoise and LeNO under single- and multi-reward optimization.}
\label{tab:training_config}
\resizebox{\linewidth}{!}{%
\begin{tabular}{lcc|cc}
\toprule
& \multicolumn{2}{c|}{\textbf{Single Reward}}
& \multicolumn{2}{c}{\textbf{Multi Reward}} \\
\cmidrule(lr){2-3}
\cmidrule(lr){4-5}
& \textbf{HyperNoise}
& \textbf{LeNO}
& \textbf{HyperNoise}
& \textbf{LeNO} \\
\midrule

Model
& SDXL-Turbo & SDXL-Turbo & SANA-Sprint & SANA-Sprint \\

Denoising Step
& 1 & 1 & 1 & 1 \\

Learning rate
& $1\times10^{-3}$
& $5\times10^{-5}$
& $1\times10^{-3}$
& $5\times10^{-5}$ \\

GradNorm Clipping
& 1.0 & 1.0 & 1.0 & 1.0 \\

LoRA rank
& 16 & 16 & 128 & 128 \\

LoRA alpha 
& 16 & 16 & 256 & 256 \\

Optimizer
& SGD & Adam & SGD & Adam \\

Batch size
& 16 & 16 & 16 & 16 \\

Accumulation steps
& 3 & 3 & 3 & 3 \\

Training steps
& 60k & 5k & 60k & 5k \\

Number of training prompts
& 55k & 55k & 55k & 55k \\

Image size
& $512\times512$ & $512\times512$ & $1024\times1024$ & $1024\times1024$ \\

\bottomrule
\end{tabular}%
}
\end{table}

% Required packages: amsmath, amssymb, booktabs, tabularx

\subsection{Protein Structure Generation}
\label{app:protein_rmf_implementation}

\paragraph{Generator and Residual Correction.}
We use the pretrained both RMF-M and RMF-L checkpoint at epoch 1600 as a frozen
one-step backbone generator. For a protein of length $N$, the initial
state consists of residue-wise translations and rotations,
$X=(T,R)\in\mathbb{R}^{N\times3}_0\times\mathrm{SO}(3)^N$,
where the translation subspace has zero center of mass.
Translations follow a centered Gaussian prior with standard deviation
$\tau_T=10\,\text{\AA}$, and rotations are sampled independently
from the Haar measure on $\mathrm{SO}(3)$.
We parameterize the residual correction using the difference between
LoRA-adapted and frozen RMF velocity fields:
\begin{equation}
(U_T,U_R)
=
f_{\theta}^{\mathrm{LoRA}}(X;0,1)-f_0(X;0,1).
\end{equation}
Rank-32 LoRA adapters with $\alpha=32$ are inserted into 284 trunk
linear layers, giving approximately 17.2M trainable parameters.
Only the adapters are optimized, and their output matrices are
initialized to zero so that the initial predicted correction is zero.

\paragraph{Stochastic Residual Transport.}
Let $\mathcal{C}$ remove the translation center of mass and
$\operatorname{clip}_c$ limit each residue-wise vector to norm $c$.
We define the translation and rotation action means as
\begin{equation}
\begin{aligned}
M_T
&=
\mathcal{C}\!\left(
\operatorname{clip}_{c_T}(\eta_T U_T)
\right),
\\
M_R
&=
\operatorname{clip}_{c_R}(\eta_R U_R),
\end{aligned}
\end{equation}
with $(\eta_T,\eta_R)=(0.30,0.15)$,
$c_T=2.0\,\text{\AA}$, and $c_R=0.10\,\text{rad}$.
The stochastic transport samples residual actions
\begin{equation}
\begin{aligned}
A_T &= M_T+\sigma_T\mathcal{C}(\epsilon_T),\\
A_R &= M_R+\sigma_R\epsilon_R,
\end{aligned}
\end{equation}
where $\epsilon_T$ and $\epsilon_R$ are independent
standard Gaussian tensors,
$\sigma_T=2.0\,\text{\AA}$, and
$\sigma_R=0.20\,\text{rad}$.
The corrected initial frames are
\begin{equation}
\begin{aligned}
T^+
&=
\|T\|_F
\frac{\mathcal{C}(T+A_T)}
{\|\mathcal{C}(T+A_T)\|_F},
\\
R_i^+
&=
R_i\operatorname{Exp}(\widehat{A}_{R,i}),
\end{aligned}
\end{equation}
where $\widehat{\cdot}$ maps a tangent vector to a
skew-symmetric matrix.
The translation update preserves the Frobenius norm of each
centered Gaussian anchor while changing its direction, and the rotation
update remains on $\mathrm{SO}(3)$.

During training, we sample multiple independent action pairs
for each anchor to construct group-relative advantages.
At inference, we sample one action pair using the same
$\sigma_T$ and $\sigma_R$ and apply a single residual
transport step before generation.

Proposition~\ref{prop:distributional_kl} extends directly to the protein
structure generation setting. The translation prior is an isotropic Gaussian
on the zero-center-of-mass subspace, and the centered noise
$\mathcal{C}(\epsilon_T)$ lies in the same subspace, so the
rotational-invariance argument applies within it. For rotations, $R_i$ is
Haar-distributed and independent of the zero-mean action, so
$R_i\operatorname{Exp}(\widehat{\sigma_R\epsilon_{R,i}})$ remains
Haar-distributed by invariance of the Haar measure. Clipping acts only on the
predicted mean and does not affect the argument.

\paragraph{Local GRPO Training.}

Each training iteration samples one initial state with
$N\sim\mathcal{U}\{60,\ldots,128\}$ and explores 36 local particles.
Rewards are standardized within each group using the population
standard deviation with $\epsilon=10^{-6}$, and advantages are
clipped to $[-5,5]$.
Exploration likelihoods are evaluated on the stored pre-projection
residuals $(A_T,A_R)$.
We perform two optimization epochs per group with surrogate clipping
$0.20$ and a Gaussian mean-shift KL regularizer toward zero residual
correction, weighted by $5\times10^{-3}$.
We use Adam with learning rate $2\times10^{-5}$,
$(\beta_1,\beta_2)=(0.9,0.999)$, $\epsilon=10^{-8}$,
no weight decay, and gradient-norm clipping at $1.0$.
We report the checkpoint at iteration 1k. The corresponding training
configurations are reported in
Tab.~\ref{tab:protein_training_config}.

\begin{table}[t]
\centering
\caption{Training configuration for protein generation.}
\label{tab:protein_training_config}
\small
\renewcommand{\arraystretch}{0.95}
\setlength{\tabcolsep}{5pt}
\begin{tabular}{@{}l@{\hspace{10pt}}l@{}}
\toprule
\textbf{Hyperparameter} & \textbf{Value} \\
\midrule
Frozen generator & RMF-M / RMF-L \\
Generator sampling steps & 1 \\
Residual transport steps & 1 \\
Training protein length & $N\sim\mathcal{U}\{60,\ldots,128\}$ \\
LoRA rank / alpha & $32/32$ \\
Trainable parameters & 7.1M / 17.2M \\
\midrule
Initial states per iteration & 1 \\
Local particles per state & 36 \\
Optimization epochs per group & 2 \\
Gradient accumulation & 1 \\
Surrogate clipping & 0.20 \\
Advantage clipping & $[-5,5]$ \\
KL coefficient & $5\times10^{-3}$ \\
\midrule
Optimizer & Adam \\
Learning rate & $2\times10^{-5}$ \\
Adam betas & $(0.9,0.999)$ \\
Weight decay & 0 \\
GradNorm clipping & 1.0 \\
Training steps & 1k \\
\midrule
Sequences per backbone (train / eval.) & $4/8$ \\
Sequence sampling temperature & 0.1 \\
Reward soft-min temperature & $1.0\,\text{\AA}$ \\
\bottomrule
\end{tabular}
\end{table}

\paragraph{Designability Reward and Evaluation.}
Each corrected initial state is decoded into a backbone by the
frozen one-step RMF generator.
ProteinMPNN samples $K=4$ sequences per backbone at temperature
$0.1$, and ESMFold (\texttt{facebook/esmfold\_v1}) refolds them.
Given their aligned self-consistency RMSDs $\{d_k\}_{k=1}^{K}$,
we maximize the negative normalized soft minimum:
\begin{equation}
\mathcal{R}
=
\tau\log\left(
\frac{1}{K}\sum_{k=1}^{K}\exp(-d_k/\tau)
\right),
\qquad \tau=1.0\,\text{\AA}.
\end{equation}
All particles within a local group share the same ProteinMPNN
seed to reduce reward variance; no pLDDT term is included.
For final evaluation, we sample eight sequences per backbone and
classify it as designable when the minimum scRMSD is below
$2\,\text{\AA}$.

\section{Additional Experiments}
\label{app:additional_experiments}

\subsection{Quality and Diversity Preservation}
\label{app:quality_diversity}

\begin{figure}[t]
    \centering
    \includegraphics[width=\textwidth]{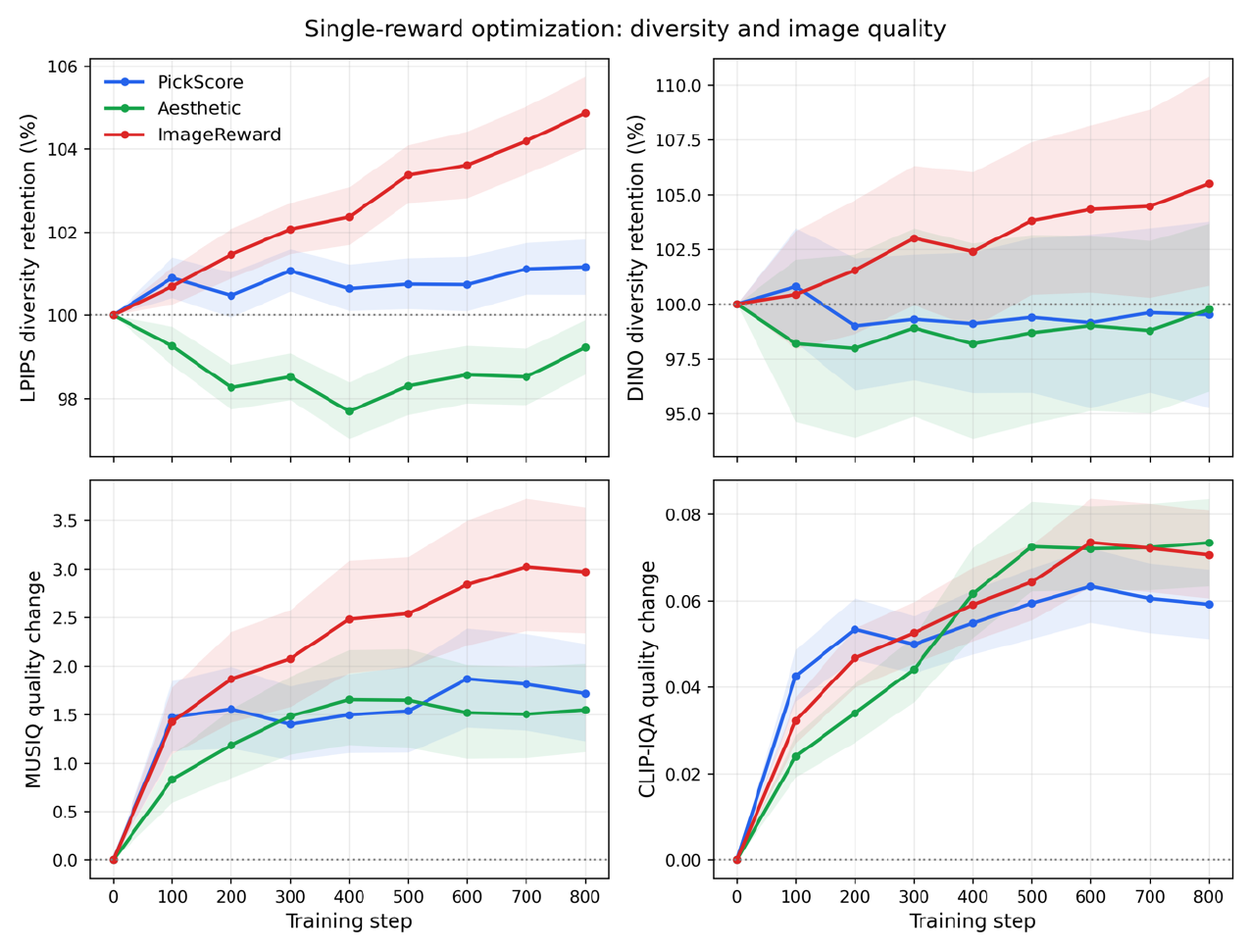}
    \caption{
        Diversity and no-reference image quality throughout single-reward
        optimization. LPIPS and DINO diversity are normalized by the
        unoptimized SDXL-Turbo baseline, while MUSIQ and CLIP-IQA are shown as
        absolute improvements over the baseline. All methods preserve nearly
        all of the baseline diversity, while consistently improving perceptual
        image quality. Shaded regions indicate 95\% paired bootstrap confidence
        intervals over prompts.
    }
    \label{fig:quality_diversity}
\end{figure}

We investigate whether optimizing the initial noise for a single reward leads
to diversity collapse or degradation in perceptual image quality. We evaluate
the PickScore-, Aesthetic-, and ImageReward-optimized models on 100 held-out
TempFlow prompts, using eight fixed initial-noise seeds per prompt. The same
prompt--seed pairs are shared across all methods and checkpoints. Since the
Aesthetic model has the shortest training horizon, we compare all models up to
800 training steps. The shaded regions in Fig.~\ref{fig:quality_diversity}
denote 95\% confidence intervals obtained by paired bootstrap resampling over
prompts.

We report two complementary diversity metrics and two no-reference image
quality metrics:
\begin{itemize}
    \item \textbf{LPIPS diversity}~\citep{zhang2018unreasonable}: the average pairwise LPIPS distance among
    the eight images generated for the same prompt. A higher value indicates
    greater perceptual variation across initial-noise samples.
    
    \item \textbf{DINO diversity}~\citep{oquab2023dinov2}: the average pairwise cosine distance between
    DINOv2 image features. This metric captures diversity at a more semantic
    and content-level representation than pixel-level distances.
    
    \item \textbf{MUSIQ}~\citep{ke2021musiq}: a no-reference image quality assessment metric that
    measures overall perceptual quality without requiring reference images.
    Higher values indicate better image quality.
    
    \item \textbf{CLIP-IQA}~\citep{wang2023exploring}: a CLIP-based no-reference image quality metric
    that assesses perceptual quality using semantic visual representations.
    Higher values indicate better quality.
\end{itemize}

For the diversity metrics, we additionally report retention relative to the
unoptimized SDXL-Turbo baseline. For the quality metrics, we report the
absolute score together with its improvement over the baseline.

\begin{table}[H]
    \centering
    \caption{
        Diversity and image quality after 800 training steps. Values in
        parentheses denote diversity retention relative to SDXL-Turbo for
        LPIPS and DINO, and absolute improvement over SDXL-Turbo for MUSIQ and
        CLIP-IQA.
    }
    \label{tab:quality_diversity}
    \resizebox{\textwidth}{!}{
    \begin{tabular}{lrrrr}
        \toprule
        Method
        & LPIPS (ret.) $\uparrow$
        & DINO div. (ret.) $\uparrow$
        & MUSIQ ($\Delta$) $\uparrow$
        & CLIP-IQA ($\Delta$) $\uparrow$ \\
        \midrule
        SDXL-Turbo
        & 0.5984 (100.0\%)
        & 0.1952 (100.0\%)
        & 70.96 (+0.00)
        & 0.6600 (+0.0000) \\
        
        PickScore (step 800)
        & 0.6054 (101.2\%)
        & 0.1942 (99.5\%)
        & 72.67 (+1.72)
        & 0.7192 (+0.0592) \\
        
        Aesthetic (step 800)
        & 0.5938 (99.2\%)
        & 0.1947 (99.7\%)
        & 72.51 (+1.55)
        & 0.7335 (+0.0734) \\
        
        ImageReward (step 800)
        & 0.6275 (104.9\%)
        & 0.2059 (105.5\%)
        & 73.93 (+2.97)
        & 0.7306 (+0.0706) \\
        \bottomrule
    \end{tabular}
    }
\end{table}

As shown in Fig.~\ref{fig:quality_diversity} and Tab.~\ref{tab:quality_diversity}, reward optimization does not
induce diversity collapse. At step 800, PickScore retains 101.2\% and 99.5\%
of the baseline LPIPS and DINO diversity, respectively. The Aesthetic model
shows only a marginal reduction, retaining 99.2\% LPIPS diversity and 99.7\%
DINO diversity. ImageReward increases both measures to 104.9\% and 105.5\%,
respectively. Across the entire optimization trajectory, the minimum observed
retention is 97.7\% for LPIPS and 98.0\% for DINO, indicating that the
seed-dependent variation is largely preserved.

At the same time, all three optimized models improve both no-reference quality
metrics. At step 800, MUSIQ increases by $+1.72$, $+1.55$, and $+2.97$ for
PickScore, Aesthetic, and ImageReward, respectively. CLIP-IQA similarly
improves by $+0.0592$, $+0.0734$, and $+0.0706$. These improvements appear
throughout training rather than only at the final checkpoint. Overall, the
results show that Local GRPO improves the target reward without sacrificing
sample diversity and simultaneously improves no-reference perceptual quality.
We note that these measurements characterize within-prompt diversity and
perceptual quality, rather than distribution-level fidelity such as FID.

\paragraph{Cross-reward evaluation.}
To further examine whether gains on the target reward come at the expense of
other preference signals, we evaluate each single-reward LeNO model on all
three rewards using the same held-out prompt--seed pairs.
As shown in Tab.~\ref{tab:cross_reward}, each model achieves its largest
improvement on its own training reward (diagonal), while the two non-target
rewards also increase in every case. For example, the PickScore-optimized model
improves ImageReward by $+0.160$, and the ImageReward-optimized model improves
both Aesthetic and PickScore. No non-target reward decreases on average,
indicating that LeNO does not exploit idiosyncrasies of a single reward model
but instead shifts generations toward outputs that are broadly preferred.

\begin{table}[t]
    \centering
    \caption{
        Cross-reward evaluation of single-reward LeNO models on SDXL-Turbo.
        Rows denote the training reward and columns the evaluation reward.
        Values in parentheses denote improvement over SDXL-Turbo; shaded
        cells indicate the target reward.
    }
    \label{tab:cross_reward}
    \begin{tabular}{lccc}
        \toprule
        \multirow{2}{*}{Training reward}
        & \multicolumn{3}{c}{Evaluation reward} \\
        \cmidrule(lr){2-4}
        & Aesthetic $\uparrow$ & PickScore $\uparrow$ & ImageReward $\uparrow$ \\
        \midrule
        SDXL-Turbo (no training)
        & 6.1174 & 21.8608 & 0.9710 \\
        \midrule
        Aesthetic
        & \cellcolor{blue!10}\textbf{6.1796} (+0.0622)
        & 21.9728 (+0.1120)
        & 1.0297 (+0.0587) \\
        PickScore
        & 6.1688 (+0.0514)
        & \cellcolor{blue!10}\textbf{22.3308} (+0.4701)
        & 1.1306 (+0.1596) \\
        ImageReward
        & 6.1558 (+0.0384)
        & 21.9092 (+0.0484)
        & \cellcolor{blue!10}\textbf{1.1984} (+0.2274) \\
        \bottomrule
    \end{tabular}
\end{table}

\subsection{Multi-step Correction Ablation}
\label{app:multistep_ablation}

We additionally investigate whether applying the learned noise correction over
multiple residual steps provides finer-grained optimization than the
single-step formulation used in the main experiments. We consider
$k\in\{1,2,4\}$ residual transport steps. For the multi-step variants, we use
a branching strategy in which stochastic perturbations are introduced at only
one of the $k$ steps, while the remaining steps follow deterministic residual
transitions. Concretely, at the selected branching step, multiple local
particles are generated around the predicted correction and evaluated using
the terminal reward; before and after this branching point, the transport is
applied deterministically. This preserves the local reward comparison used by
Local GRPO without exponentially branching the trajectory at every step.

\begin{figure}[t]
    \centering
    \includegraphics[width=0.8\linewidth]{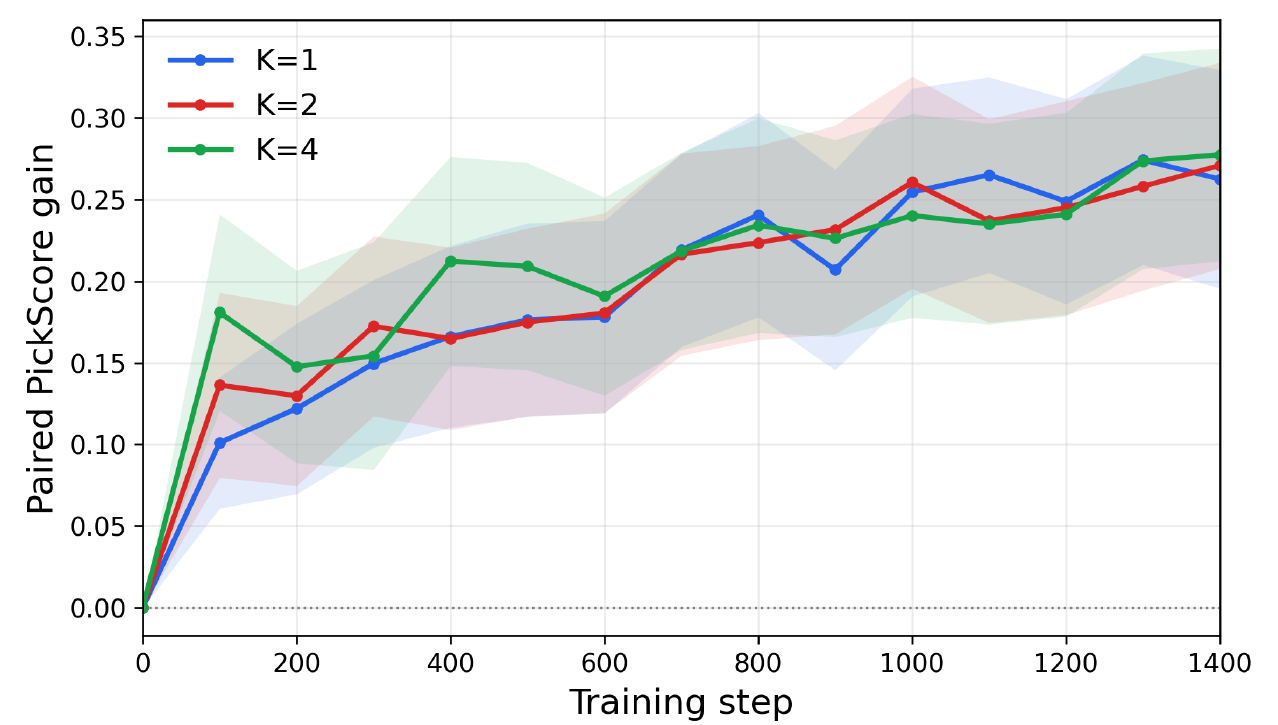}
    \vspace{-1mm}
    \caption{Ablation for multi-step transport reward gain.}
    \label{fig:multistep_ablation}
\end{figure}

Figure~\ref{fig:multistep_ablation} compares the reward gain throughout
training for $k=1,2,$ and $4$. All three settings exhibit nearly identical
improvement curves, and increasing the number of residual steps does not lead
to a clear additional gain. Although a multi-step transport is in principle
more expressive and could progressively refine the corrected noise, we do not
observe such a benefit in this setting. One possible explanation is that the
single-step state-dependent correction already captures most of the useful
local reward-improvement direction. Moreover, under the branching strategy,
only one step receives stochastic local exploration in each branched
trajectory, while the remaining steps compose deterministic transformations. Therefore, increasing $k$ does not necessarily provide additional independent
reward-improvement signals. These results suggest that iterative refinement is
not essential for the tasks considered here, motivating the use of the simpler
single-step transport in our main experiments.

\subsection{KL Regularization Ablation}
\label{app:kl_ablation}

We study the effect of the KL regularization weight
$\beta$, which controls how strongly the learned residual
transport is constrained to remain close to the reference noise distribution.
We evaluate
$\beta\in\{0,0.01,0.1,1.0\}$ using ImageReward as the
optimization objective.

\begin{figure}[t]
\centering

\begin{minipage}[t]{0.58\linewidth}
\vspace{0pt}
\centering
\includegraphics[width=\linewidth]{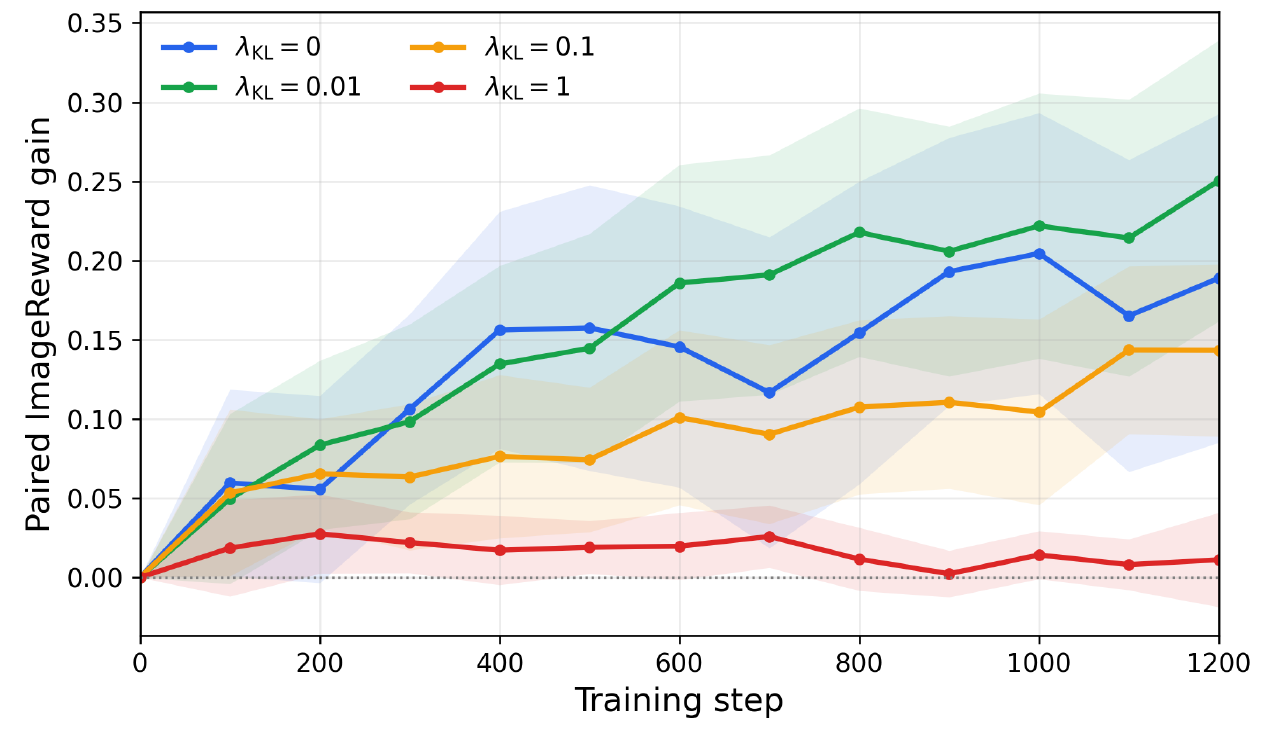}
\vspace{-5mm}
\captionof{figure}{
Ablation of the KL regularization weight $\beta$.
}
\label{fig:kl_ablation}
\end{minipage}%
\hfill
\begin{minipage}[t]{0.38\linewidth}
\vspace{5mm}
\centering
\captionof{table}{
Effect of $\beta$ on ImageReward gain and win rate.
}
\vspace{-2mm}
\label{tab:kl_ablation}
\resizebox{\linewidth}{!}{%
\begin{tabular}{ccc}
\toprule
$\beta$
& ImageReward Gain
& Win Rate \\
\midrule
$0$
& $0.1889$
& $69\%$ \\
$0.01$
& $\mathbf{0.2504}$
& $\mathbf{76\%}$ \\
$0.1$
& $0.1433$
& $68\%$ \\
$1.0$
& $0.0112$
& $59\%$ \\
\bottomrule
\end{tabular}%
}
\end{minipage}

\end{figure}

As shown in Fig.~\ref{fig:kl_ablation}, a small amount of KL regularization
improves both the final reward gain and training stability.
Without KL regularization ($\beta=0$), LeNO still improves
the reward substantially, but achieves a lower final gain
($+0.1889\pm0.1038$) and win rate ($69\%$) than
$\beta=0.01$, which obtains the best result
($+0.2504\pm0.0887$, $76\%$ win rate).
Increasing the regularization further reduces the attainable improvement:
$\beta=0.1$ yields a gain of
$+0.1433\pm0.0543$, while $\beta=1.0$ nearly suppresses
the reward improvement entirely.

These results suggest that the KL term plays more than a purely
reward-regularization role. When its weight is zero, the residual transport
can move the corrected-noise distribution more freely away from the reference
prior, which may weaken the local structure assumed by the residual
parameterization and lead to less stable optimization. A small KL weight acts
as a trust-region constraint, allowing sufficient movement toward
high-reward regions while preventing excessively large distributional shifts.
Conversely, overly strong regularization restricts the correction magnitude
and prevents the transport from reaching high-reward noise states.
Accordingly, an intermediate KL weight provides the best trade-off between
reward improvement and preservation of the reference noise structure.
This behavior is consistent with the reward-tilted target, which balances
reward maximization against deviation from the reference noise distribution
rather than optimizing reward alone.

\subsection{Generalization to a One-Step GAN Generator}
\label{app:dfgan}

We further evaluate whether LeNO is applicable beyond diffusion-based
generators by applying it to DF-GAN~\citep{tao2022df}, a one-step prompt-conditioned GAN. We use the same LeNO training setup as in the main experiments, keeping the pretrained generator frozen and optimizing only the residual correction from scalar reward evaluations. Since LeNO operates entirely in the initial noise space and does not require gradients through either the generator or the reward function, no modification of the generator architecture or its
internal sampling process is required.

\begin{figure}[t]
    \centering
    \includegraphics[width=\linewidth]{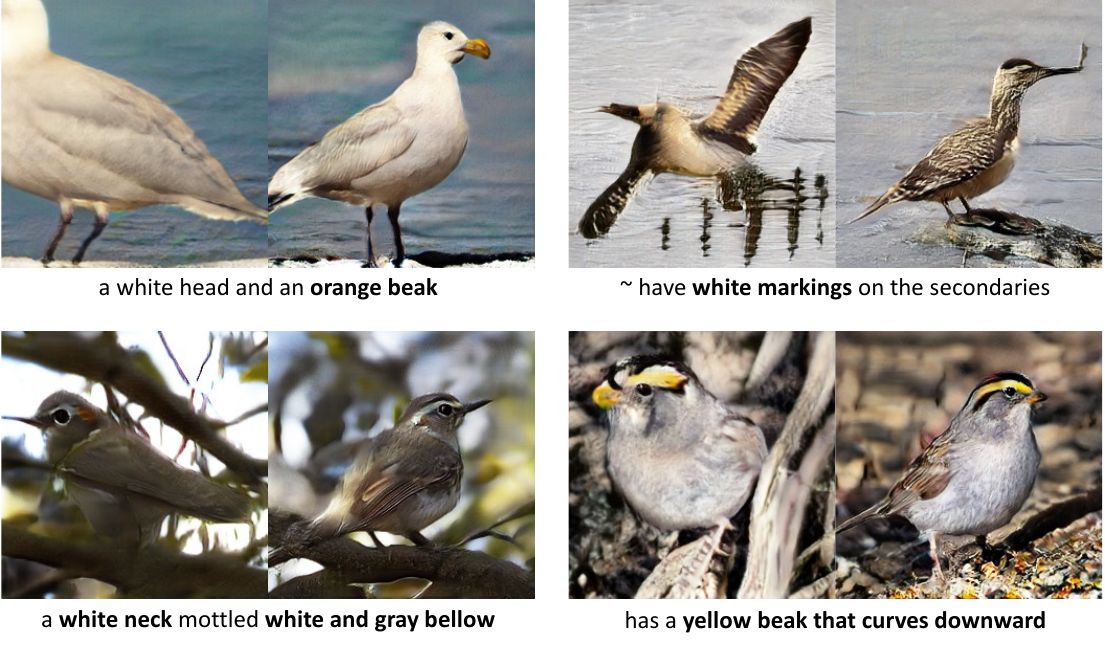}
    \caption{Qualitative comparison of DF-GAN generated samples before (left) and after LeNO noise correction (right). }
    \label{fig:dfgan_qualitative}
\end{figure}

Figure~\ref{fig:dfgan_qualitative} shows qualitative examples before and after
LeNO correction to PickScore. Across different prompts, the corrected initial noises
produce samples with improved reward-aligned visual characteristics while
preserving the underlying generation capability of the frozen DF-GAN.
Importantly, this experiment uses the same optimization procedure despite the
substantially different generator family. The result therefore supports that
LeNO is not tied to a particular diffusion formulation or denoising
trajectory, but can be applied to deterministic prompt-conditioned generators
whenever they expose an initial noise input and can be evaluated by a scalar
reward oracle. This provides additional evidence for the black-box
applicability of LeNO to both the generator and the reward function.

\section{Additional Results}
\label{app:additional_qualitative_results}

\begin{figure}[t]
    \centering
    \includegraphics[width=\linewidth]{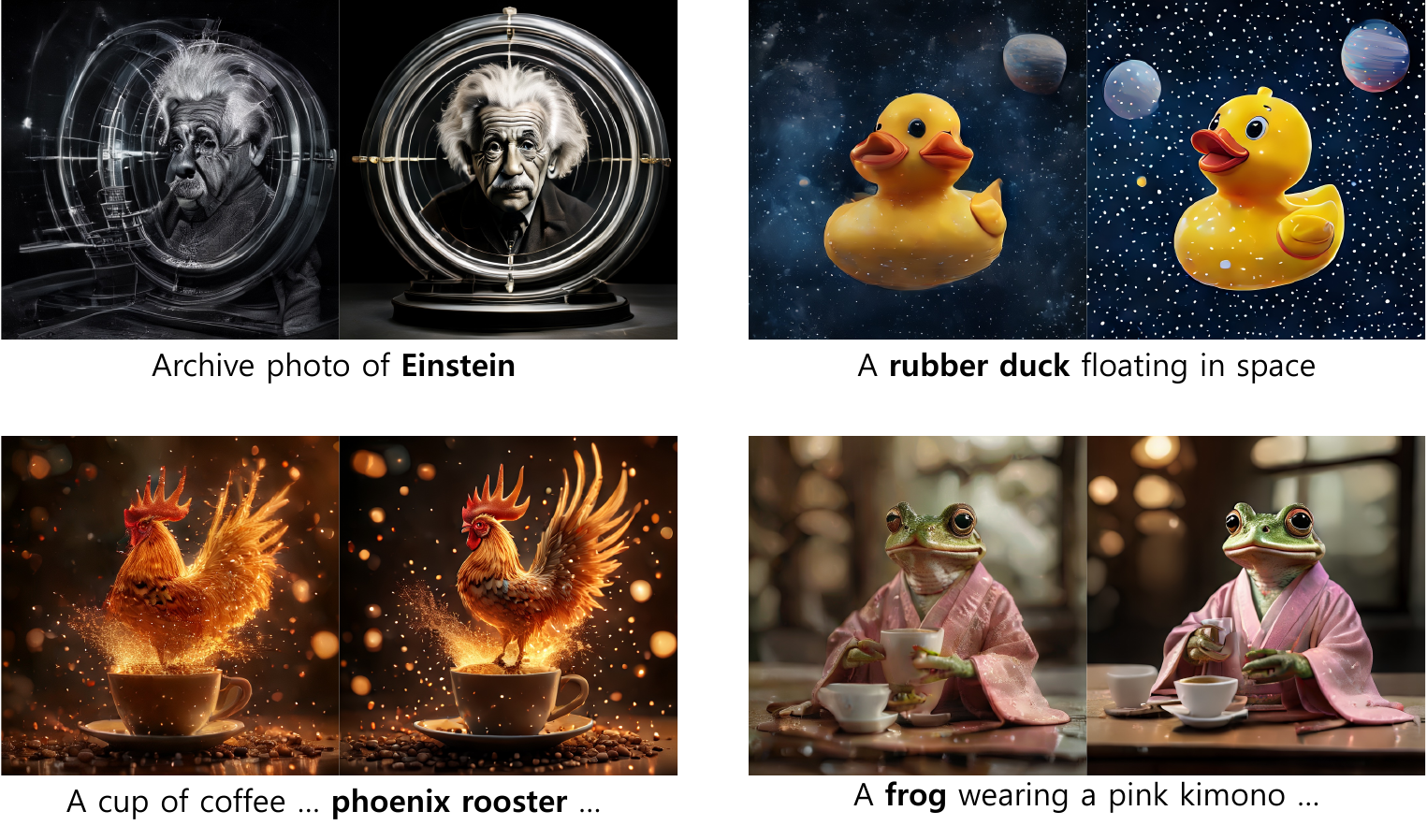}
    \caption{Additional qualitative comparison of SANA-Sprint optimized with compositional multi-rewards.}
    \label{fig:additional_SANA}
\end{figure}

\begin{figure}[t]
    \centering
    \includegraphics[width=\linewidth]{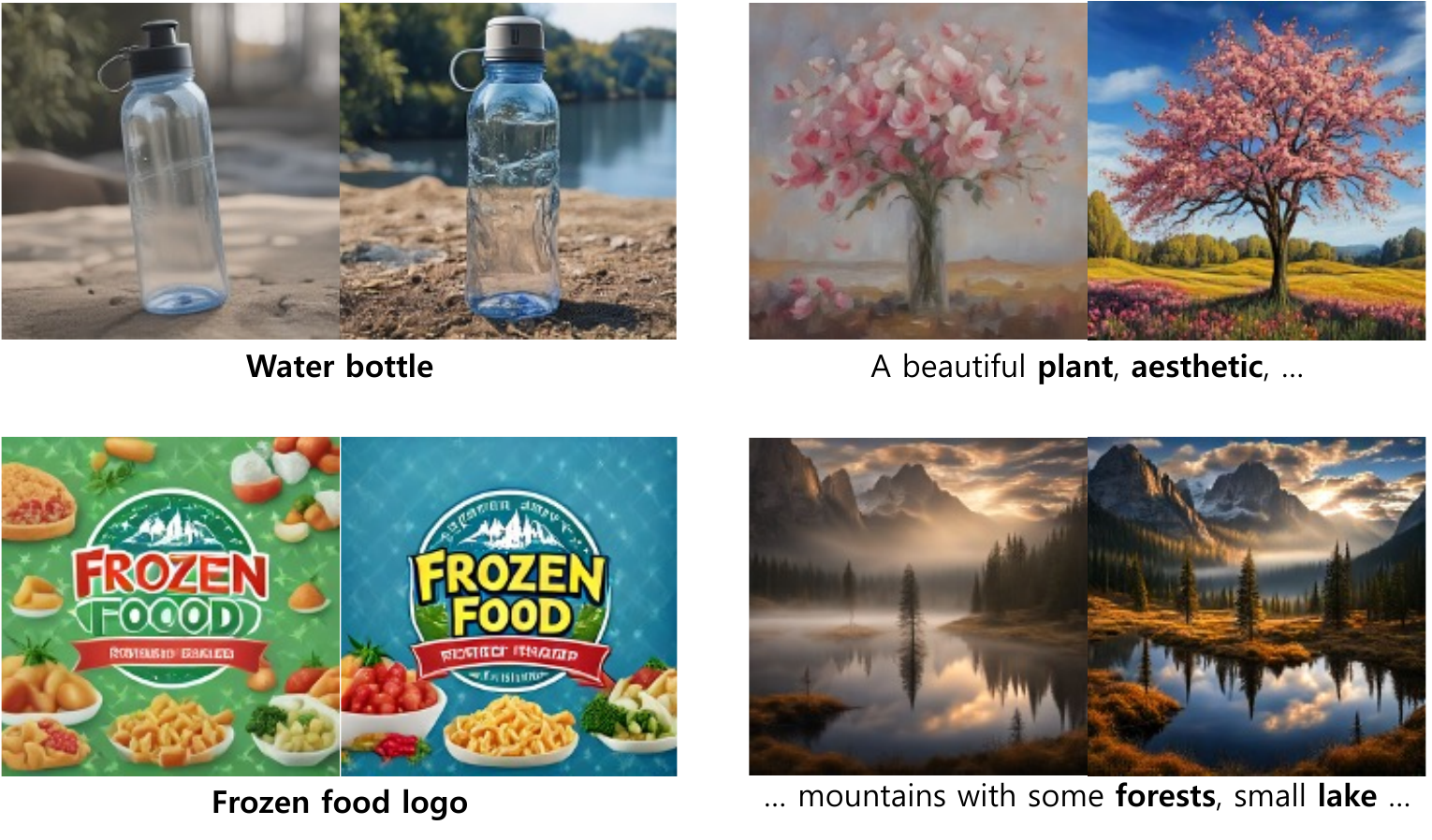}
    \caption{Additional qualitative comparison of SDXL-Turbo optimized with Image Rewards.}
    \label{fig:additional_SDXL_Img}
\end{figure}

\begin{figure}[t]
    \centering
    \includegraphics[width=\linewidth]{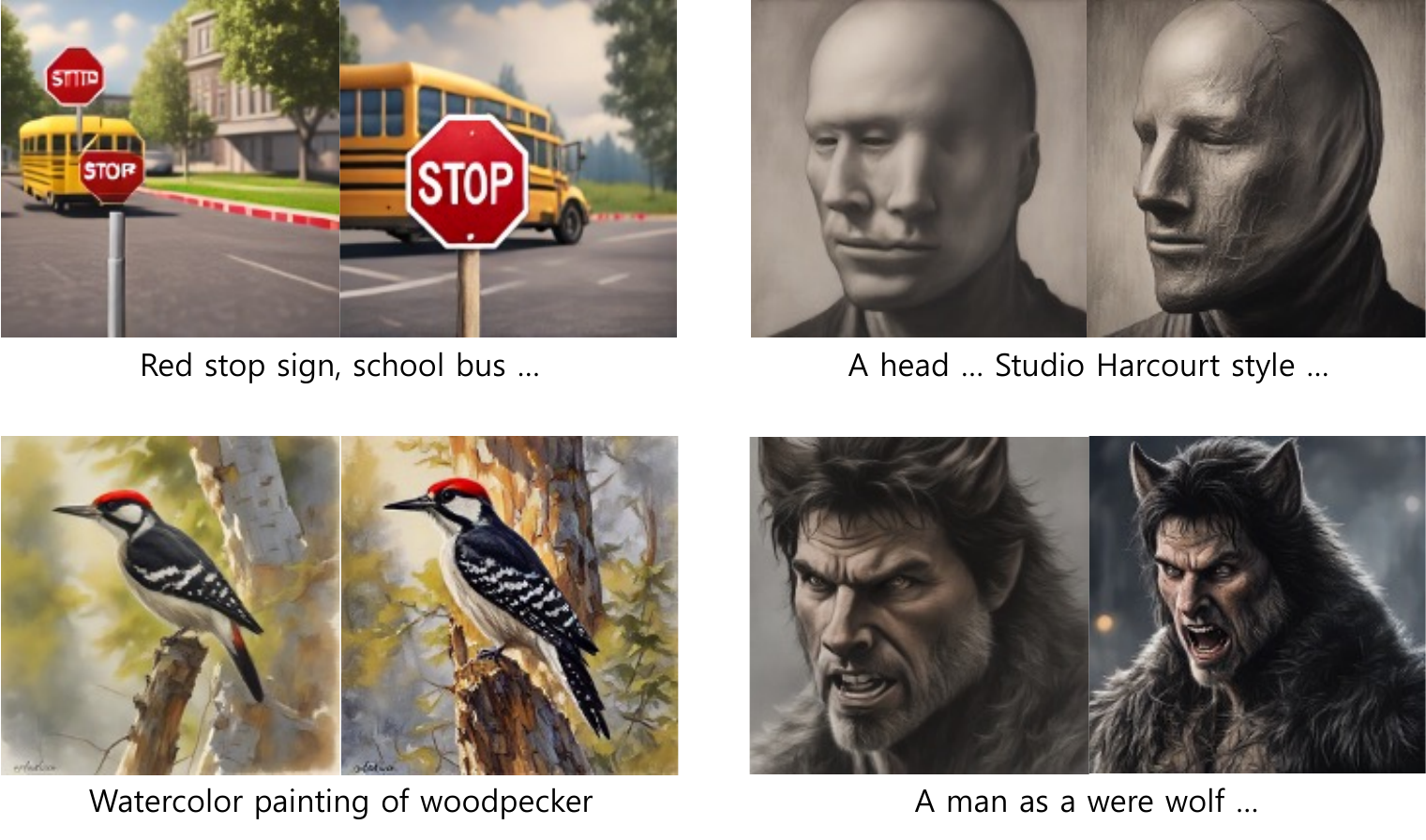}
    \caption{Additional qualitative comparison of SDXL-Turbo optimized with PickScore.}
    \label{fig:additional_SDXL_Pick}
\end{figure}

\begin{figure}[t]
    \centering
    \includegraphics[width=\linewidth]{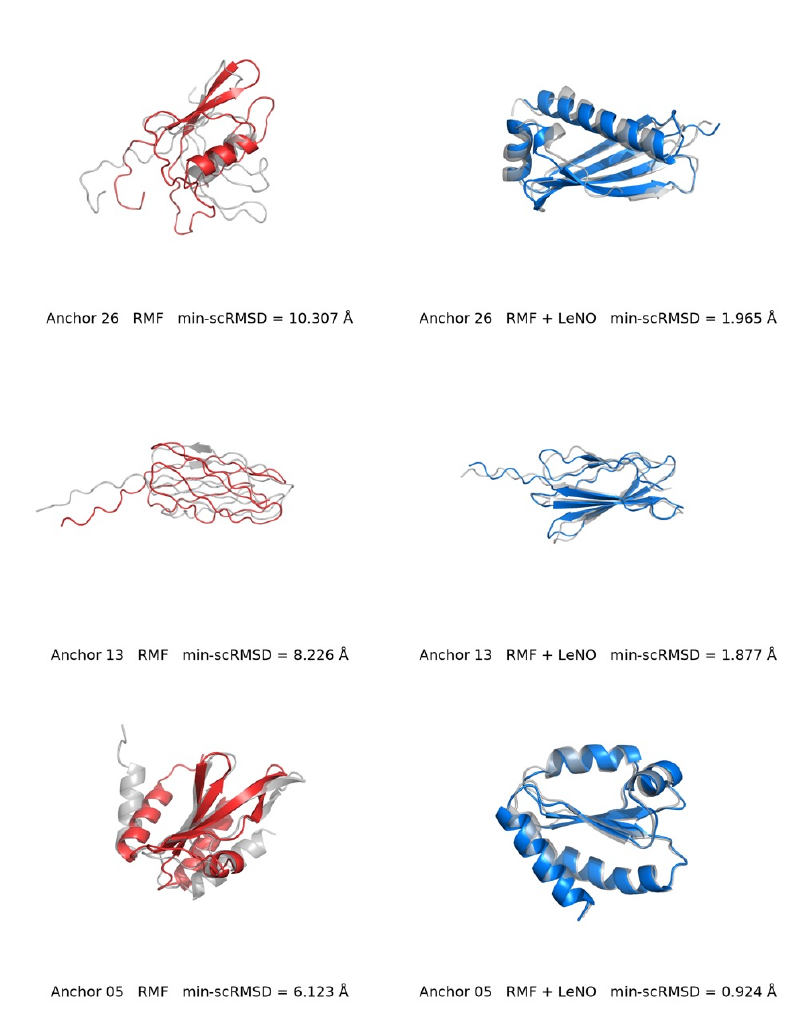}
    \caption{Additional qualitative self-consistency comparison with RMF at protein length $L$=96.}
    \label{fig:additional_protein}
\end{figure}

\end{document}